\documentclass[10pt,twocolumn,letterpaper]{article}

\usepackage{cvpr}              %

\usepackage{graphicx}
\usepackage{amsmath}
\usepackage{amssymb}
\usepackage{booktabs}
\usepackage[accsupp]{axessibility} 

\usepackage{pifont}

\newcommand{\xmark}{\ding{55}}

\usepackage{times}
\usepackage{epsfig}
\usepackage{url}
\usepackage{xcolor}
\usepackage{multirow}
\usepackage{wrapfig}
\usepackage{graphicx}
\usepackage{subcaption}
\usepackage{booktabs} %
\newcommand{\ra}[1]{\renewcommand{\arraystretch}{#1}} %
\usepackage{multirow} %
\usepackage{url} %
\usepackage{cuted} %
\usepackage{capt-of}
\usepackage[font={small}]{caption}
\usepackage{epigraph}
\usepackage{enumitem}
\usepackage{xspace}
\usepackage{censor}

\DeclareMathOperator*{\argmin}{argmin}
\newcommand{\norm}[1]{\left\lVert#1\right\rVert}

\newcommand{\dset}{ABO\xspace} %
\newcommand{\dsetlong}{Amazon Berkeley Objects\xspace}

\newcommand{\nlistings}{147,702 } %
\newcommand{\nprodtypes}{576 } %
\newcommand{\ncategoriesthreed}{63 } %
\newcommand{\nposedthreed}{6,334 } %
\newcommand{\nthreed}{7,953 } %
\newcommand{\nspins}{8,222 } %
\newcommand{\nimages}{398,212 } %

\usepackage[pagebackref,breaklinks,colorlinks]{hyperref}

\usepackage[capitalize]{cleveref}
\crefname{section}{Sec.}{Secs.}
\Crefname{section}{Section}{Sections}
\Crefname{table}{Table}{Tables}
\crefname{table}{Tab.}{Tabs.}

\def\cvprPaperID{6061\vspace{-30pt}} %
\def\confName{CVPR}
\def\confYear{2022}

\begin{document}

\title{\dset: Dataset and Benchmarks for Real-World 3D Object Understanding
\vspace*{-2mm}
}

\author{
Jasmine Collins\textsuperscript{1}, Shubham Goel\textsuperscript{1}, Kenan Deng\textsuperscript{2}, Achleshwar Luthra\textsuperscript{3}, Leon Xu\textsuperscript{1,2}, \\ 
Erhan Gundogdu\textsuperscript{2}, Xi Zhang\textsuperscript{2}, Tomas F. Yago Vicente\textsuperscript{2}, Thomas Dideriksen\textsuperscript{2}, \\
Himanshu Arora\textsuperscript{2}, Matthieu Guillaumin\textsuperscript{2}, and Jitendra Malik\textsuperscript{1} \\ \\
\textsuperscript{1} UC Berkeley, \textsuperscript{2} Amazon, \textsuperscript{3} BITS Pilani
\vspace*{-0.2in}
}

\maketitle
\begin{strip}\centering
\includegraphics[width=0.95\textwidth]{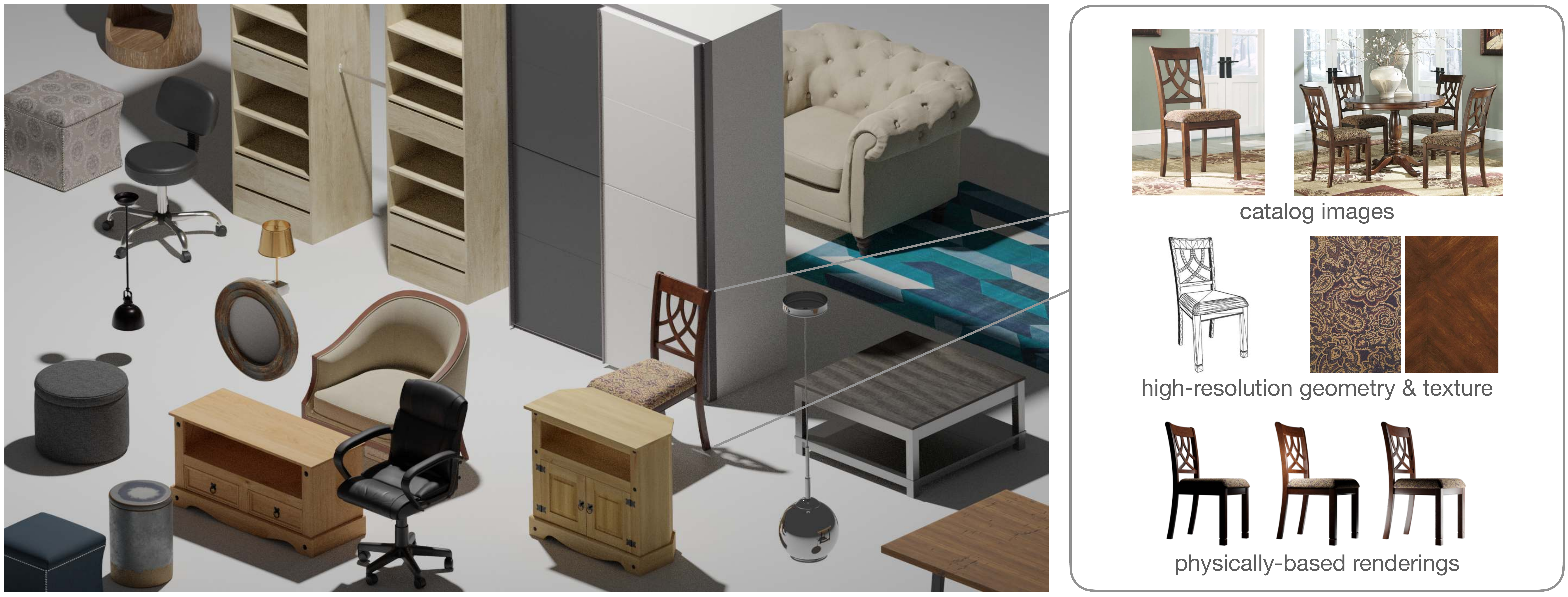}
\vspace*{-2mm}
\captionof{figure}{{\bf \dset is a dataset of product images and realistic, high-resolution, physically-based 3D models of household objects.} We use \dset to benchmark the performance of state-of-the-art methods on a variety of realistic object understanding tasks.}
\label{fig:teaser}
\end{strip}

\begin{abstract}
\vspace*{-2mm}
We introduce \dsetlong (\dset), a new large-scale dataset designed to help bridge the gap between real and virtual 3D worlds. \dset contains product catalog images, metadata, and artist-created 3D models with complex geometries and physically-based materials that correspond to real, household objects. We derive challenging benchmarks that exploit the unique properties of \dset and measure the current limits of the state-of-the-art on three open problems for real-world 3D object understanding: single-view 3D reconstruction, material estimation, and cross-domain multi-view object retrieval.

\vspace{-15pt}
\end{abstract}
\section{Introduction}
Progress in 2D image recognition has been driven by large-scale datasets~\cite{krizhevsky2009learning, deng2009imagenet, lin2014microsoft, sun2017revisiting, gupta2019lvis}. The ease of collecting 2D annotations (such as class labels or segmentation masks) has led to the large scale of these diverse, in-the-wild datasets, which in turn has enabled the development of 2D computer vision systems that work in the real world. Theoretically, progress in 3D computer vision should follow from equally large-scale datasets of 3D objects.
However, collecting large amounts of high-quality 3D annotations (such as voxels or meshes) for individual real-world objects poses a challenge. 
One way around the challenging problem of getting 3D annotations for real images is to focus only on synthetic, computer-aided design (CAD) models~\cite{chang2015shapenet, zhou2016thingi10k, koch2019abc}. This has the advantage that the data is large in scale (as there are many 3D CAD models available for download online) but many of the models are low quality or untextured and do not exist in the real world. 
This has led to a variety of 3D reconstruction methods that work well on clear-background renderings of synthetic objects~\cite{choy20163d, groueix2018papier, xie2019pix2vox, mescheder2019occupancy} but do not necessarily generalize to real images, new categories, or more complex object geometries~\cite{tatarchenko2019single, bautista2021generalization, bechtold2021fostering}. 

To enable better real-world transfer, another class of 3D datasets aims to link existing 3D models with real-world images~\cite{xiang2014pascal, xiang2016objectnet3d}. These datasets find the closest matching CAD models for the objects in an image and have human annotators align the pose of each model to best match the image. While this has enabled the evaluation of 3D reconstruction methods in-the-wild, the shape (and thus pose) matches are approximate. Further, because this approach relies on matching CAD models to images, it inherits the limitations of the existing CAD model datasets (\ie poor coverage of real-world objects, basic geometries and %
textures).

The IKEA~\cite{lim2013parsing} and Pix3D~\cite{sun2018pix3d} datasets sought to improve upon this by annotating real images with exact, pixel-aligned 3D models. The exact nature of such datasets has allowed them to be used as training data for single-view reconstruction~\cite{meshrcnn} and has bridged some of the synthetic-to-real domain gap. However, the size of the datasets are relatively small (90 and 395 unique 3D models, respectively), likely due to the difficulty of finding images that exactly match 3D models.
Further, the larger of the two datasets~\cite{sun2018pix3d} only contains 9 categories of objects. The provided 3D models are also untextured, thus the annotations in these datasets are typically used for shape or pose-based tasks, rather than tasks such as material prediction.

Rather than trying to match images to synthetic 3D models, another approach to collecting 3D datasets is to start with real images (or video) and reconstruct the scene by classical reconstruction techniques such as structure from motion, multi-view stereo and texture mapping~\cite{choi2016large, singh2014bigbird, googleobj}. The benefit of these methods is that the reconstructed geometry faithfully represents an object of the real world. However, the collection process requires a great deal of manual effort and thus datasets of this nature tend to also be quite small (398, 125, and 1032 unique 3D models, respectively). The objects are also typically imaged in a controlled lab setting and do not have corresponding real images of the object ``in context''. Further, included textured surfaces are assumed to be Lambertian and thus do not display realistic reflectance properties.

\begin{figure}[t]
    \centering
    \includegraphics[width=.92\linewidth]{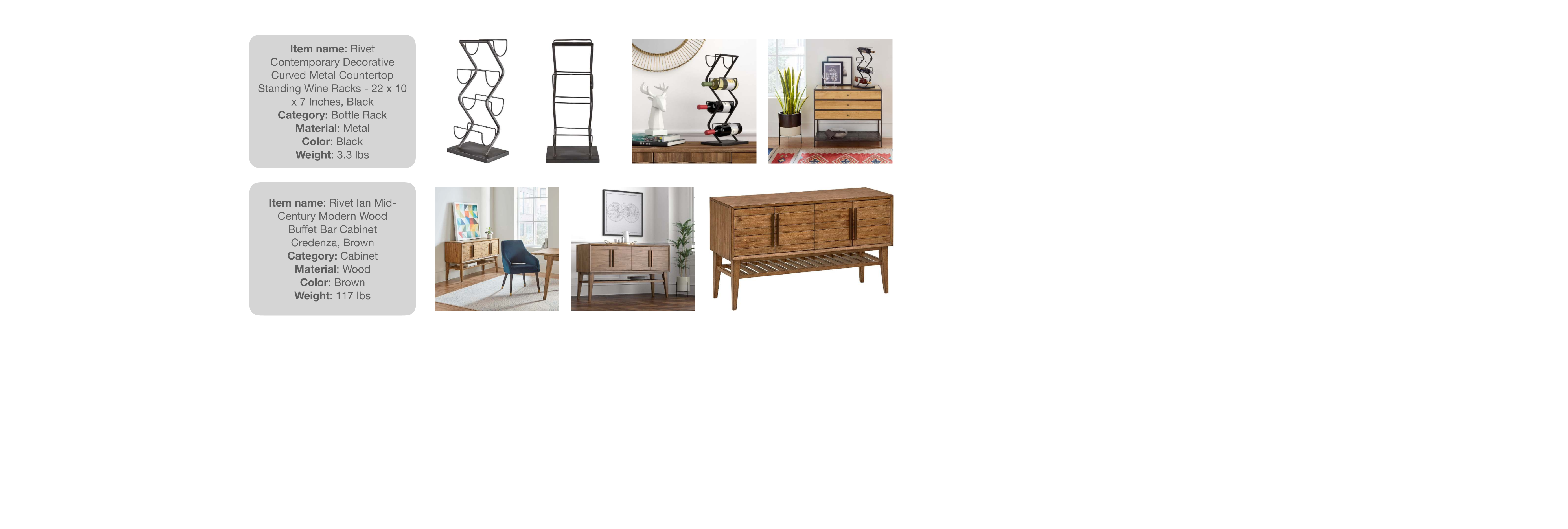}
    \caption{\textbf{Sample catalog images and attributes that accompany \dset objects.} Each object has up to 18 attribute annotations.}
    \label{fig:abo_attr}
    \vspace{10pt}
    \includegraphics[width=.92\linewidth]{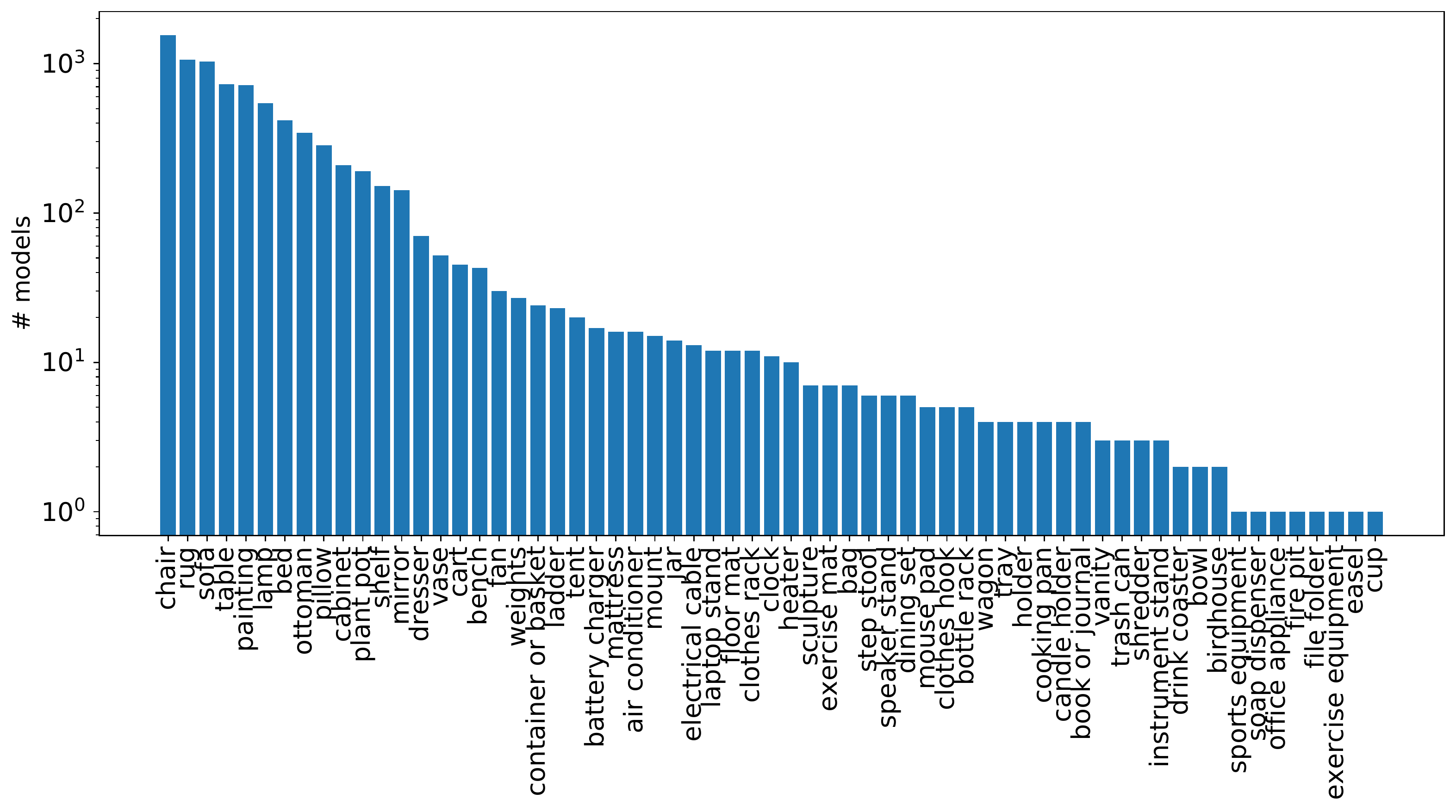}
    \caption{\textbf{3D model categories.} Each category is also mapped to a synset in the WordNet hierarchy. Note the y-axis is in log scale.}
    \label{fig:abo_hist}
    \vspace{-5pt}
\end{figure}

Motivated by the lack of large-scale datasets with realistic 3D objects from a diverse set of categories and corresponding real-world multi-view images, we introduce \dsetlong (\dset). This dataset is derived from Amazon.com product listings, and as a result, contains imagery and 3D models that correspond to modern, real-world, household items.
Overall, \dset contains \nlistings product listings associated with \nimages unique catalog images, and up to 18 unique metadata attributes (category, color, material, weight, dimensions, etc.) per product. \dset also includes ``360º View'' turntable-style images for $\nspins$ products and \nthreed products with corresponding artist-designed 3D meshes.
In contrast to existing 3D computer vision datasets, the 3D models in \dset have complex geometries and high-resolution, physically-based materials that allow for photorealistic rendering. 
A sample of the kinds of real-world images associated with a 3D model from \dset can be found in Figure~\ref{fig:teaser}, and sample metadata attributes are shown in Figure~\ref{fig:abo_attr}. The dataset is released under CC BY-NC 4.0 license and can be downloaded at {\small\url{ https://amazon-berkeley-objects.s3.amazonaws.com/index.html}}.

To facilitate future research, we benchmark the performance of various methods on three computer vision tasks that can benefit from more realistic 3D datasets: (i) single-view shape reconstruction, where we measure the domain gap for networks trained on synthetic objects, (ii) material estimation, where we introduce a baseline for spatially-varying BRDF from single- and multi-view images of complex real world objects, and (iii) image-based multi-view object retrieval, where we leverage the 3D nature of \dset to evaluate the robustness of deep metric learning algorithms to object viewpoint and scenes.

\begin{table}[t]
\begin{center}
{
    \setlength\tabcolsep{3pt}\footnotesize
    \ra{1.1} %
\begin{tabular}{lccccc}
  \toprule
  Dataset & \# Models & \# Classes & Real images & Full 3D & PBR \\
  \midrule
  ShapeNet~\cite{chang2015shapenet} & 51.3K & 55 & \xmark & \checkmark &  \xmark \\
    3D-Future~\cite{fu20203d} & 
  16.6K
  & 8 & \xmark & \checkmark & \xmark \\

  Google Scans~\cite{googleobj} & 
  1K
  & - & \xmark & \checkmark & \xmark \\
  
  CO3D~\cite{reizenstein2021common} & 
  18.6K
  & 50 & \checkmark & \xmark & \xmark \\
  IKEA~\cite{lim2013ikea} & 219 & 11 & \checkmark & \checkmark & \xmark \\
    
  Pix3D~\cite{sun2018pix3d} & 395 & 9 & \checkmark & \checkmark & \xmark \\
  
    PhotoShape~\cite{park2018photoshape} & 
  5.8K
  & 1 & \xmark & \checkmark & \checkmark \\ %
  
  \textbf{\dset (Ours)} & 
  8K
  & \ncategoriesthreed & \checkmark & \checkmark & \checkmark \\
  \bottomrule
\end{tabular}
}
\end{center}
\vspace*{-5mm}
\caption{\textbf{A comparison of the 3D models in \dset and other commonly used object-centric 3D datasets.} \dset contains nearly 8K 3D models with physically-based rendering (PBR) materials and corresponding real-world catalog images.
}
\vspace{-5pt}
\label{tab:dataset_comparison}
\end{table}

\begin{figure}[t]
    \centering
    \includegraphics[width=1.0\linewidth]{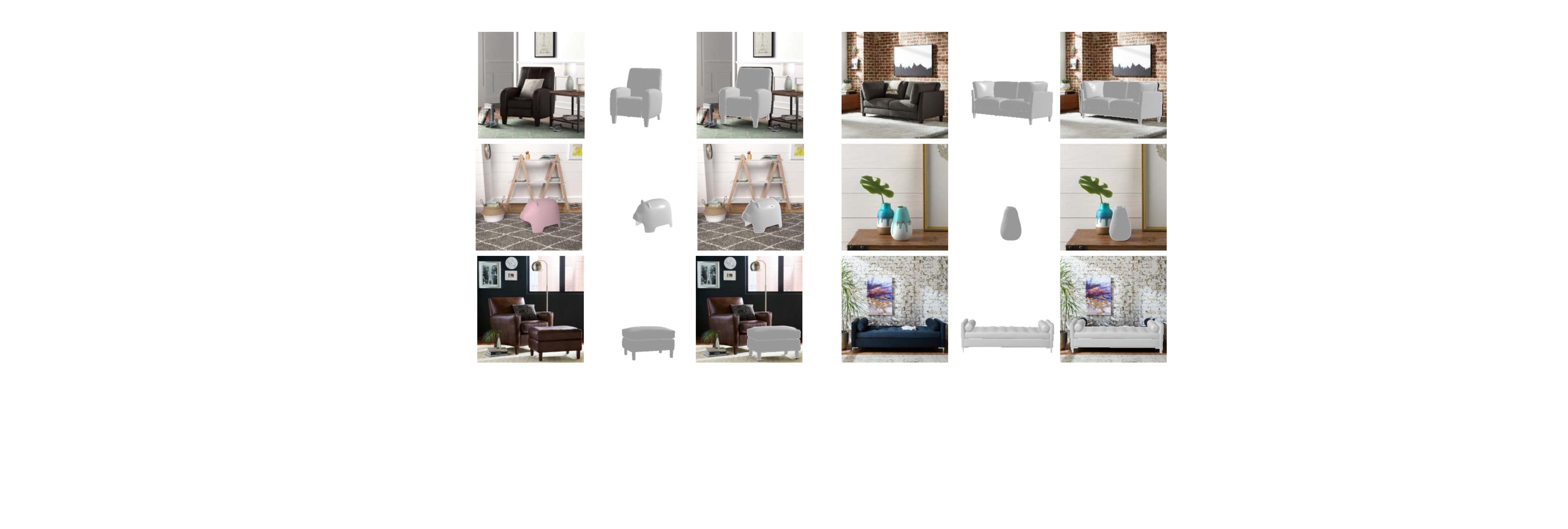}
    
    \caption{\textbf{Posed 3D models in catalog images.} We use instance masks to automatically generate 6-DOF pose annotations.}
    \vspace{-5pt}
    \label{fig:posed}
\end{figure}

\begin{table*}[t]
\begin{center}
{\footnotesize
\setlength\tabcolsep{3pt}
    \ra{1.1} %
\begin{tabular}{llrrrrcrrrrlrl}
    \toprule

  \multirow{2}{*}{Benchmark} &
  \multirow{2}{*}{Domain} &
  \multirow{2}{*}{Classes} &
  \multicolumn{3}{c}{Instances} & &
  \multicolumn{4}{c}{Images} &
  \multirow{2}{*}{Structure} &
  \multirow{2}{*}{Recall@1} & \\ %
  \cline{4-6} \cline{8-11}
  & & & train & val & test & & train & val & test-target & test-query & &  & %
  \\
  \midrule
CUB-200-2011 &
  Birds &
  200 &
  - &
  - &
  - & &
  5994 &
  0 &
  - &
  5794 &
  15 parts &
  79.2\% & \cite{jun2019combination} \\
Cars-196 &
  Cars &
  196 &
  - &
  - &
  - & &
  8144 &
  0 &
  - &
  8041 &
  - &
  94.8\% & \cite{jun2019combination} \\
In-Shop &
  Clothes &
  25 &
  3997 &
  0 &
  3985 & &
  25882 &
  0 &
  12612 &
  14218 &
  Landmarks, poses, masks &
  92.6\% & \cite{Kim_2020_CVPR} \\
SOP &
  Ebay &
  12 &
  11318 &
  0 &
  11316 & &
  59551 &
  0 &
  - &
  60502 &
  - &
  84.2\% & \cite{jun2019combination} \\
  \dset (MVR) &
  Amazon &
  562 &
  49066 &
  854 &
  836 & &
  298840 &
  26235 &
  4313 &
  23328 &
  Subset with 3D models &
  30.0\% \\ %
    \bottomrule
\end{tabular}
}
\end{center}
\vspace*{-5mm}
\caption{\textbf{Common image retrieval benchmarks for deep metric learning and their statistics.} Our proposed multi-view retrieval (MVR) benchmark based on \dset is significantly larger, more diverse and challenging than existing benchmarks, and exploits 3D models.}
\vspace{-5pt}
\label{tab:retrieval_benchmarks}
\end{table*}

\section{Related Work}
\label{sec:related}
\noindent \textbf{3D Object Datasets}
ShapeNet~\cite{chang2015shapenet} is a large-scale database of synthetic 3D CAD models commonly used for training single- and multi-view reconstruction models. 
IKEA Objects~\cite{lim2013ikea} and Pix3D~\cite{sun2018pix3d} are image collections with 2D-3D alignment between CAD models and real images, however these images are limited to objects for which there is an exact CAD model match. Similarly, Pascal3D+~\cite{xiang2014pascal} and ObjectNet3D~\cite{xiang2016objectnet3d} provide 2D-3D alignment for images and provide more instances and categories, however the 3D annotations are only approximate matches. 
The Object Scans dataset~\cite{choi2016large} and Objectron~\cite{ahmadyan2020objectron} are both video datasets that have the camera operator walk around various objects, but are limited in the number of categories represented. CO3D~\cite{reizenstein2021common} also offers videos of common objects from 50 different categories, however they do not provide full 3D mesh reconstructions. 

Existing 3D datasets typically assume very simplistic texture models that are not physically realistic. To improve on this, PhotoShapes \cite{park2018photoshape} augmented ShapeNet CAD models by automatically mapping spatially varying (SV-) bidirectional reflectance distribution functions (BRDFs) to meshes, yet the dataset consists only of chairs. The works in~\cite{deschaintre2019flexible,gao2019deep} provide high-quality SV-BRDF maps, but only for planar surfaces. The dataset used in~\cite{kim2017lightweight} contains only homogenous BRDFs for various objects. \cite{li2018learning} and~\cite{bi2020deep} introduce datasets containing full SV-BRDFs, however their models are procedurally generated shapes that do not correspond to real objects. In contrast, \dset provides shapes and SV-BRDFs created by professional artists for real-life objects that can be directly used for photorealistic rendering.

Table \ref{tab:dataset_comparison} compares the 3D subset of \dset with other commonly used 3D datasets in terms of size (number of objects and classes) and properties such as the presence of real images, full 3D meshes and physically-based rendering (PBR) materials. \dset is the only dataset that contains all of these properties and is much more diverse in number of categories than existing 3D datasets.

\medskip
\noindent \textbf{3D Shape Reconstruction}
Recent methods for single-view 3D reconstruction differ mainly in the type of supervision and 3D representation used, whether it be voxels, point clouds, meshes, or implicit functions. Methods that require full shape supervision in the single-view~\cite{fan2017point, zhang2018genre, sun2018pix3d, mescheder2019occupancy, gkioxari2019mesh} and multi-view~\cite{kar2017learning, choy20163d, xie2019pix2vox} case are often trained using ShapeNet. There are other approaches that use more natural forms of multi-view supervision such as images, depth maps, and silhouettes~\cite{yan2016perspective, wiles2017silnet, kar2017learning, tulsiani2017multi}, with known cameras. Of course, multi-view 3D reconstruction has long been studied with classical computer vision techniques~\cite{hartley2003multiple} like multi-view stereo and visual hull reconstruction. Learning-based methods are typically trained in a category-specific way and evaluated on new instances from the same category. Out of the works mentioned, only~\cite{zhang2018genre} claims to be category-agnostic. In this work we are interested in how well these ShapeNet-trained networks~\cite{choy20163d,zhang2018genre,mescheder2019occupancy,gkioxari2019mesh} generalize to more realistic objects.

\begin{figure*}[h]
     \centering
     \includegraphics[width=0.9\textwidth]{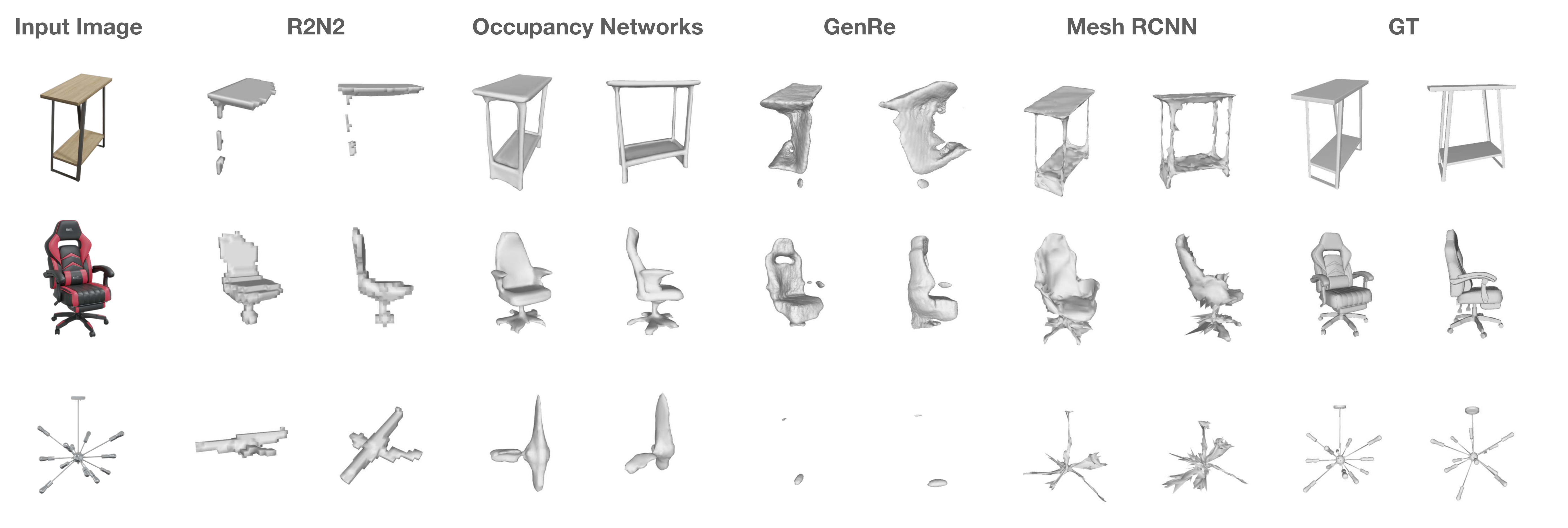}
   \vspace{-5pt}
   \caption{\textbf{Qualitative 3D reconstruction results for R2N2, Occupancy Networks, GenRe, and Mesh-RCNN on \dset.} All methods are pre-trained on ShapeNet and show a decrease in performance on objects from \dset.}
\label{fig:3d-qual}
\end{figure*}

\medskip
\noindent \textbf{Material Estimation}
Several works have focused on modeling object appearance from a single image, however realistic datasets available for this task are relatively scarce and small in size. 
\cite{li2017modeling} use two networks to estimate a homogeneous BRDF and an SV-BRDF of a flat surface from a single image, using a self-augmentation scheme to alleviate the need for a large training set. However, their work is limited to a specific family of materials, and each separate material requires another trained network. \cite{ye2018single} extend the idea of self-augmentation to train with unlabeled data, but their work is limited by the same constraints. \cite{deschaintre2018single} use a modified U-Net and rendering loss to predict the SV-BRDFs of flash-lit photographs consisting of only a flat surface. To enable prediction for arbitrary shapes, \cite{li2018learning} propose a cascaded CNN architecture with a single encoder and separate decoder for each SV-BRDF parameter. 
While the method achieves good results on semi-uncontrolled lighting environments, it requires using the intermediate bounces of global illumination rendering as supervision. 
More recent works have turned towards using multiple images to improve SV-BRDF estimation, but still only with simplistic object geometries. For instance, \cite{deschaintre2019flexible} and~\cite{gao2019deep} use multiple input images with a flash lit light source, but only for a single planar surface. \cite{bi2020deep} and~\cite{boss2020two} both use procedurally generated shapes to estimate SV-BRDFs from multi-view images. 
\dset addresses the lack of sufficient realistic data for material estimation, and in this work we propose a simple baseline method that can estimate materials from single or multi-view images of complex, real-world shapes.

\medskip
\noindent \textbf{2D/3D Image Retrieval}
Learning to represent 3D shapes and natural images of products in a single embedding space has been tackled by~\cite{li2015jointembedding}. 
They consider various relevant tasks, including cross-view image retrieval, shape-based image retrieval and image-based shape retrieval, but all are inherently constrained by the limitations of ShapeNet~\cite{chang2015shapenet} (cross-view image retrieval is only considered for chairs and cars). \cite{KrauseStarkDengFei-Fei_3DRR2013} introduced 3D object representations for fine-grained recognition and a dataset of cars with real-world 2D imagery (CARS-196), which is now widely used for deep metric learning (DML) evaluation. Likewise, other datasets for DML focus on instances/fine categories of few object types, such as birds~\cite{WahCUB_200_2011}, clothes~\cite{liuLQWTcvpr16DeepFashion}, or a few object categories~\cite{Song_2016_CVPR}.

Due to the limited diversity and the similar nature of query and target images in existing retrieval benchmarks, the performance of state-of-the-art DML algorithms are near saturation. Moreover, since these datasets come with little structure, the opportunities to analyze failure cases and improve algorithms are limited.
Motivated by this, we derive a challenging large-scale benchmark dataset from \dset with hundreds of diverse categories and a proper validation set. We also leverage the 3D nature of \dset to measure and improve the robustness of representations with respect to changes in viewpoint and scene.
A comparison of \dset and existing benchmarks for DML can be found in Table~\ref{tab:retrieval_benchmarks}.

\section{The \dset Dataset}
\label{sec:dataset}

\noindent \textbf{Dataset Properties}
The \dset dataset originates from worldwide product listings, metadata, images and 3D models provided by Amazon.com.
This data consists of \nlistings listings of  
products from \nprodtypes product types sold by various Amazon-owned stores and websites (e.g. Amazon, PrimeNow, Whole Foods).
Each listing is identified by an item ID and is provided with structured metadata corresponding to information that is publicly available on the listing's main webpage (such as product type, material, color, and dimensions) as well as the media available for that product.
This includes $\nimages$ high-resolution catalog images, and,
when available, the turntable images that are used for the ``360º View'' feature that shows the product imaged at 5º or 15º azimuth intervals ($\nspins$ products).

\begin{table*}[t]\setlength\tabcolsep{3.5pt}\small
    \centering
    \ra{1.3} %
    \begin{tabular}{@{}lccccccccccccc@{}}
    \toprule
    & \multicolumn{6}{c}{Chamfer Distance ($\downarrow$)}  && \multicolumn{6}{c}{Absolute Normal Consistency ($\uparrow$)}
    \\
    \cmidrule{2-7} \cmidrule{9-14}
     & \footnotesize{bench} & \footnotesize{chair} & \footnotesize{couch} & \footnotesize{cabinet} & \footnotesize{lamp} & \footnotesize{table} 
     && 
     \footnotesize{bench} & \footnotesize{chair} & \footnotesize{couch} & \footnotesize{cabinet} & \footnotesize{lamp} & \footnotesize{table} 
     \\
     \midrule
    {3D R2N2 \cite{choy20163d}}
    &   \scriptsize{2.46{\color{gray}/0.85}} 
    &   \scriptsize{1.46{\color{gray}/0.77}} 
    &   \scriptsize{1.15{\color{gray}/0.59}} 
    &   \scriptsize{1.88{\color{gray}/0.25}} 
    &   \scriptsize{3.79{\color{gray}/2.02}} 
    &   \scriptsize{2.83{\color{gray}/0.66}} 
    &&  \scriptsize{0.51{\color{gray}/0.55}} 
    &   \scriptsize{0.59{\color{gray}/0.61}} 
    &   \scriptsize{0.57{\color{gray}/0.62}} 
    &   \scriptsize{0.53{\color{gray}/0.67}} 
    &   \scriptsize{0.51{\color{gray}/0.54}} 
    &   \scriptsize{0.51{\color{gray}/0.65}} 
    \\
    {Occ Nets \cite{mescheder2019occupancy}} 
    &   \scriptsize{1.72{\color{gray}/0.51}} 
    &   \scriptsize{0.72{\color{gray}/0.39}} 
    &   \scriptsize{0.86{\color{gray}/0.30}} 
    &   \scriptsize{0.80{\color{gray}/0.23}} 
    &   \scriptsize{2.53{\color{gray}/1.66}} 
    &   \scriptsize{1.79{\color{gray}/0.41}} 
    &&  \scriptsize{0.66{\color{gray}/0.68}} 
    &   \scriptsize{0.67{\color{gray}/0.76}} 
    &   \scriptsize{0.70{\color{gray}/0.77}} 
    &   \scriptsize{0.71{\color{gray}/0.77}} 
    &   \scriptsize{0.65{\color{gray}/0.69}} 
    &   \scriptsize{0.67{\color{gray}/0.78}} 
    \\
    {GenRe \cite{zhang2018genre}} 
    &   \scriptsize{1.54{\color{gray}/2.86}} 
    &   \scriptsize{0.89{\color{gray}/0.79}} 
    &   \scriptsize{1.08{\color{gray}/2.18}} 
    &   \scriptsize{1.40{\color{gray}/2.03}} 
    &   \scriptsize{3.72{\color{gray}/2.47}} 
    &   \scriptsize{2.26{\color{gray}/2.37}} 
    &&  \scriptsize{0.63{\color{gray}/0.56}} 
    &   \scriptsize{0.69{\color{gray}/0.67}} 
    &   \scriptsize{0.66{\color{gray}/0.60}} 
    &   \scriptsize{0.62{\color{gray}/0.59}} 
    &   \scriptsize{0.59{\color{gray}/0.57}} 
    &   \scriptsize{0.61{\color{gray}/0.59}} 
    \\
    {Mesh R-CNN \cite{gkioxari2019mesh} }
    &   \scriptsize{1.05{\color{gray}/0.09}} 
    &   \scriptsize{0.78{\color{gray}/0.13}} 
    &   \scriptsize{0.45{\color{gray}/0.10}} 
    &   \scriptsize{0.80{\color{gray}/0.11}} 
    &   \scriptsize{1.97{\color{gray}/0.24}} 
    &   \scriptsize{1.15{\color{gray}/0.12}} 
    &&  \scriptsize{0.62{\color{gray}/0.65}} 
    &   \scriptsize{0.62{\color{gray}/0.70}} 
    &   \scriptsize{0.62{\color{gray}/0.72}} 
    &   \scriptsize{0.65{\color{gray}/0.74}} 
    &   \scriptsize{0.57{\color{gray}/0.66}} 
    &   \scriptsize{0.62{\color{gray}/0.74}} 
    \\
    \bottomrule
    \end{tabular}
    \caption{\textbf{Single-view 3D reconstruction generalization from ShapeNet to \dset.} Chamfer distance and absolute normal consistency of predictions made on \dset objects from common ShapeNet classes. We report the same metrics for ShapeNet objects (denoted in {\color{gray} gray}), following the same evaluation protocol. All methods, with the exception of GenRe, are trained on all of the ShapeNet categories listed.
    }
    \vspace{-5pt}
    \label{tab:3drecon}
\end{table*}

\medskip
\noindent \textbf{3D Models}
\dset also includes $\nthreed$ artist-created high-quality 3D models in glTF 2.0 format.
The 3D models are oriented in a canonical coordinate system where the ``front'' (when well defined) of all objects are aligned and each have a scale corresponding to real world units. To enable these meshes to easily be used for comparison with existing methods trained on 3D datasets such as ShapeNet, we have collected category annotations for each 3D model and mapped them to noun synsets under the WordNet~\cite{miller1995wordnet} taxonomy. Figure~\ref{fig:abo_hist} shows a histogram of the 3D model categories. 

\medskip
\noindent \textbf{Catalog Image Pose Annotations}
We additionally provide 6-DOF pose annotations for $\nposedthreed$ of the catalog images. To achieve this, we develop an automated pipeline for pose estimation based on the knowledge of the 3D model in the image, off-the-shelf instance masks~\cite{maskRCNN,kirillov2020pointrend}, and differentiable rendering. For each mask $\textbf{M}$, we estimate $\textbf{R} \in SO(3)$ and $\textbf{T} \in \mathbb{R}^3$ 
such that the following silhouette loss is minimized

$$\textbf{R}^*, \textbf{T}^* = \argmin_{\textbf{R},\textbf{T}} \norm{DR(\textbf{R},\textbf{T}) - \textbf{M}} $$

where $DR(\cdot)$ is a differentiable renderer implemented in PyTorch3D~\cite{ravi2020accelerating}. Examples of results from this approach can be found in Figure~\ref{fig:posed}. Unlike previous approaches to CAD-to-image alignment~\cite{xiang2016objectnet3d,sun2018pix3d} that use human annotators in-the-loop to provide pose or correspondences, our approach is fully automatic except for a final human verification step. 

\medskip
\noindent \textbf{Material Estimation Dataset}
To perform material estimation from images, we 
use the Disney~\cite{burley2012physically} base color, metallic, roughness parameterization given in glTF 2.0 specification~\cite{gltf}. We render 512x512 images from 91 camera positions along an upper icosphere of the object with a $60^\circ$ field-of-view using Blender's~\cite{blender} Cycles path-tracer. To ensure diverse realistic lighting conditions and backgrounds, we illuminate the scene using 3 random environment maps out of 108 indoor HDRIs \cite{HDRIHaven}. For these rendered images, we generate the corresponding ground truth base color, metallicness, roughness, and normal maps along with the object depth map and segmentation mask. The resulting dataset consists of 2.1 million rendered images and corresponding camera intrinsics and extrinsics.

\section{Experiments}

\subsection{Evaluating Single-View 3D Reconstruction}
\label{sec:3d}
As existing methods are largely trained in a fully supervised manner using ShapeNet~\cite{chang2015shapenet}, we are interested in how well they will transfer to more real-world objects. To measure how well these models transfer to real object instances, we evaluate the performance of a variety of these methods on objects from \dset. Specifically we evaluate 3D-R2N2~\cite{choy20163d}, GenRe~\cite{zhang2018genre}, Occupancy Networks~\cite{mescheder2019occupancy}, and Mesh R-CNN~\cite{gkioxari2019mesh} pre-trained on ShapeNet. We selected these methods because they capture some of the top-performing single-view 3D reconstruction methods from the past few years and are varied in the type of 3D representation that they use (voxels in~\cite{choy20163d}, spherical maps in~\cite{zhang2018genre}, implicit functions in~\cite{mescheder2019occupancy}, and meshes in~\cite{gkioxari2019mesh}) and the coordinate system used (canonical vs. view-space).
While all the models we consider are pre-trained on ShapeNet, GenRe trains on a different set of classes and takes as input a silhouette mask at train and test time.

To study this question (irrespective of the question of cross-category generalization), we consider only the subset of \dset models objects that fall into ShapeNet training categories. Out of the \ncategoriesthreed categories in \dset with 3D models, we consider 6 classes that intersect with commonly used ShapeNet classes, capturing 4,170 of the \nthreed 3D models. Some common ShapeNet classes, such as ``airplane'', have no matching \dset category; similarly, some categories in \dset like ``air conditioner'' and ``weights'' do not map well to ShapeNet classes. 

For this experiment, we render a dataset (distinct from the \dset Material Estimation Dataset) of objects on a blank background from a similar distribution of viewpoints as in the rendered ShapeNet training set. We render 30 viewpoints of each mesh using Blender \cite{blender},
each with a $40^\circ$ field-of-view and such that the entire object is visible. Camera azimuth and elevation are sampled uniformly on the surface of a unit sphere with a $-10^\circ$ lower limit on elevations to avoid uncommon bottom views. 

GenRe and Mesh-RCNN make their predictions in ``view-space" (i.e. pose aligned to the image view), whereas R2N2 and Occupancy Networks perform predictions in canonical space (predictions are made in the same category-specific, canonical pose despite the pose of the object in an image). For each method we evaluate Chamfer Distance and Absolute Normal Consistency and largely follow the evaluation protocol of~\cite{gkioxari2019mesh}.

\medskip
\noindent \textbf{Results}
A quantitative comparison of the four methods we considered on \dset objects can be found in Table~\ref{tab:3drecon}. 
We also re-evaluated each method's predictions on the ShapeNet test set from R2N2~\cite{choy20163d} with our evaluation protocol and report those metrics. We observe that Mesh R-CNN~\cite{gkioxari2019mesh} outperforms all other methods across the board on both \dset and ShapeNet in terms of Chamfer Distance, whereas Occupancy Networks performs the best in terms of Absolute Normal Consistency. As can be seen, there is a large performance gap between all ShapeNet and ABO predictions.
This suggests that shapes and textures from \dset, while derived from the same categories but from the real world, are out of distribution and more challenging for the models trained on ShapeNet. Further, we notice that the \textit{lamp} category has a particularly large performance drop from ShapeNet to \dset. Qualitative results suggest that this is likely due to the difficulty in reconstructing thin structures. We highlight some qualitative results in Figure~\ref{fig:3d-qual}, including one particularly challenging lamp instance.

\begin{figure*}[h]
	\centering
	\includegraphics[width=0.45\linewidth]{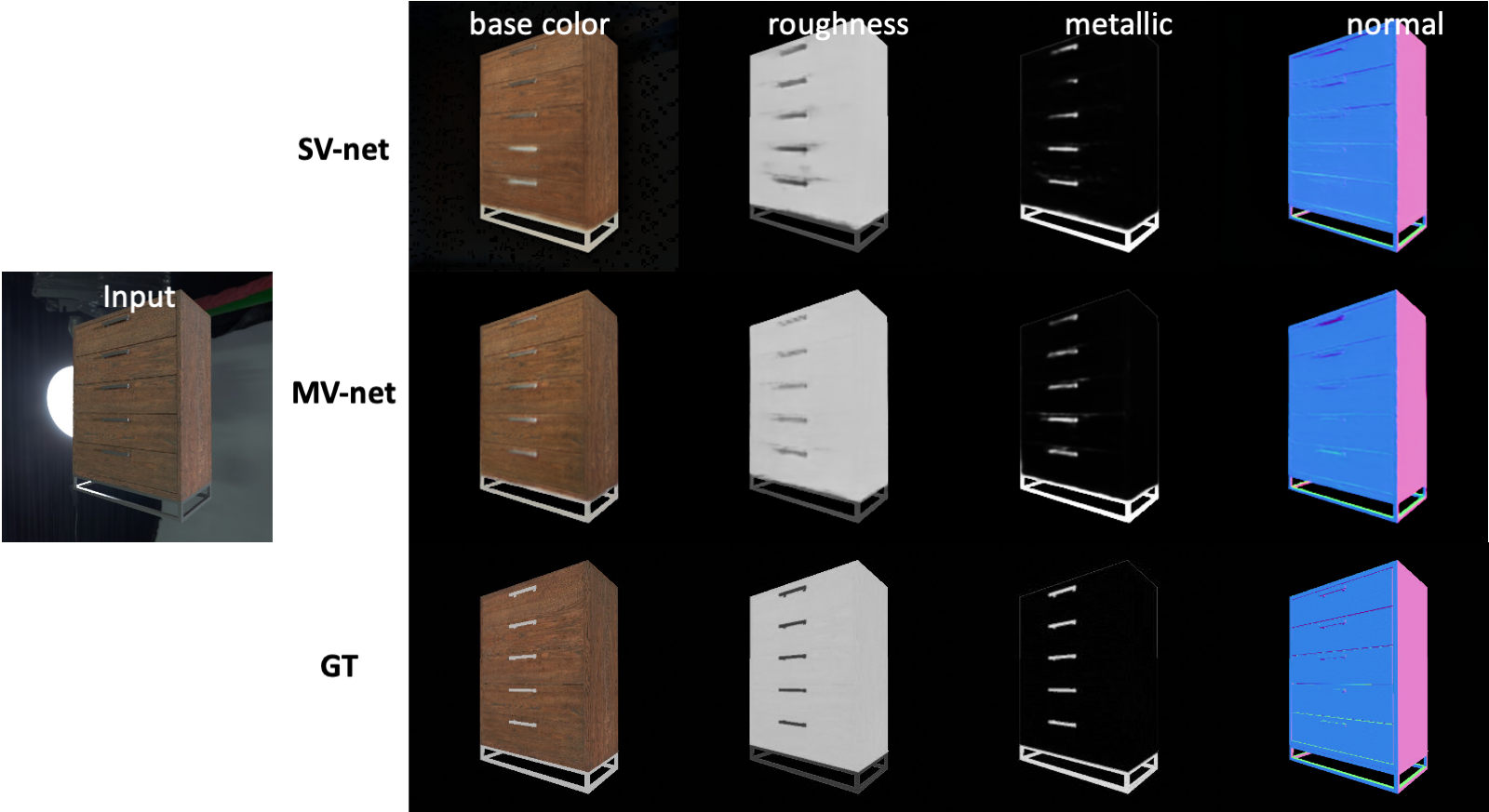}
	\includegraphics[width=0.45\linewidth]{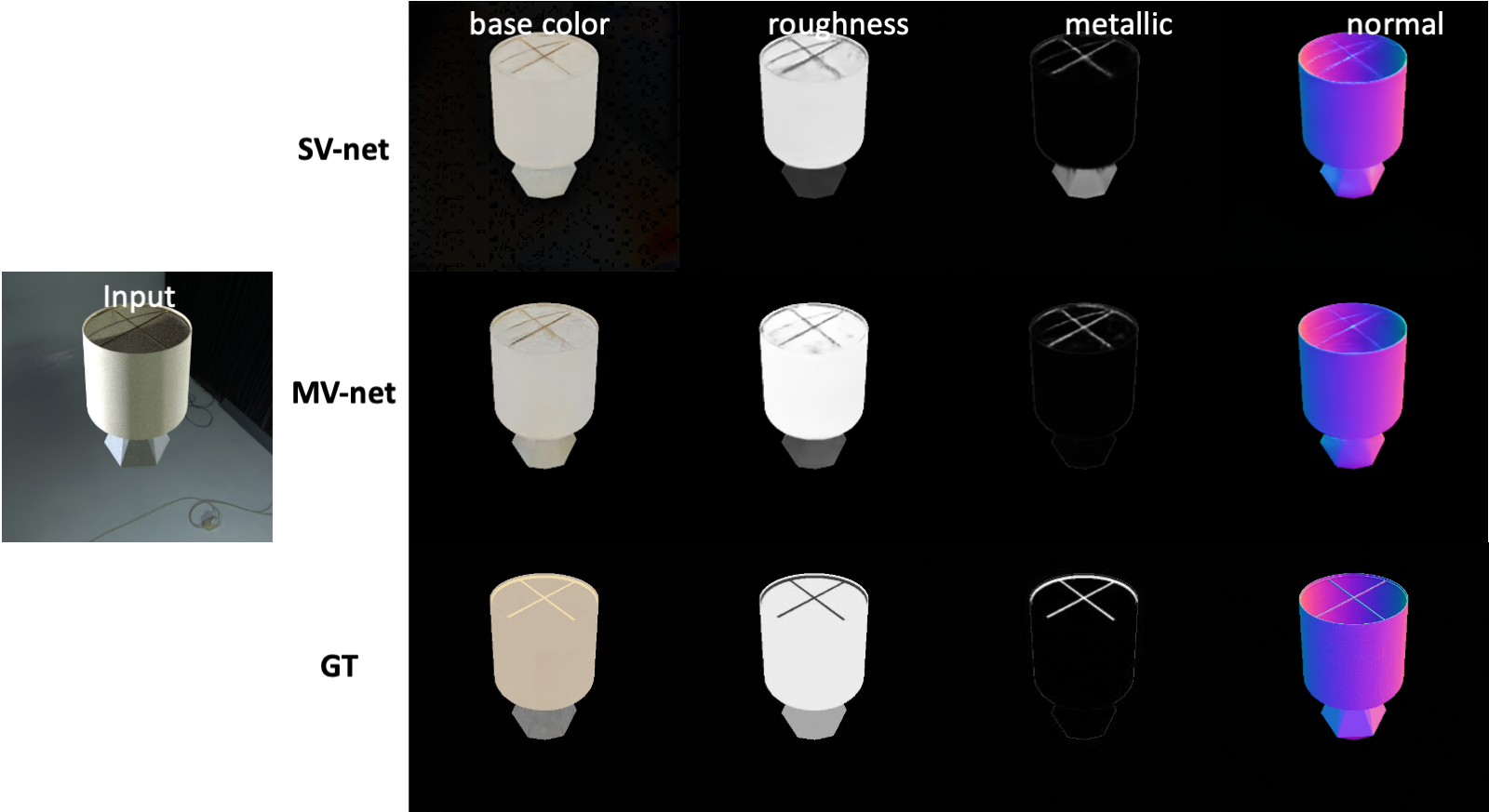}
	\includegraphics[width=0.45\linewidth]{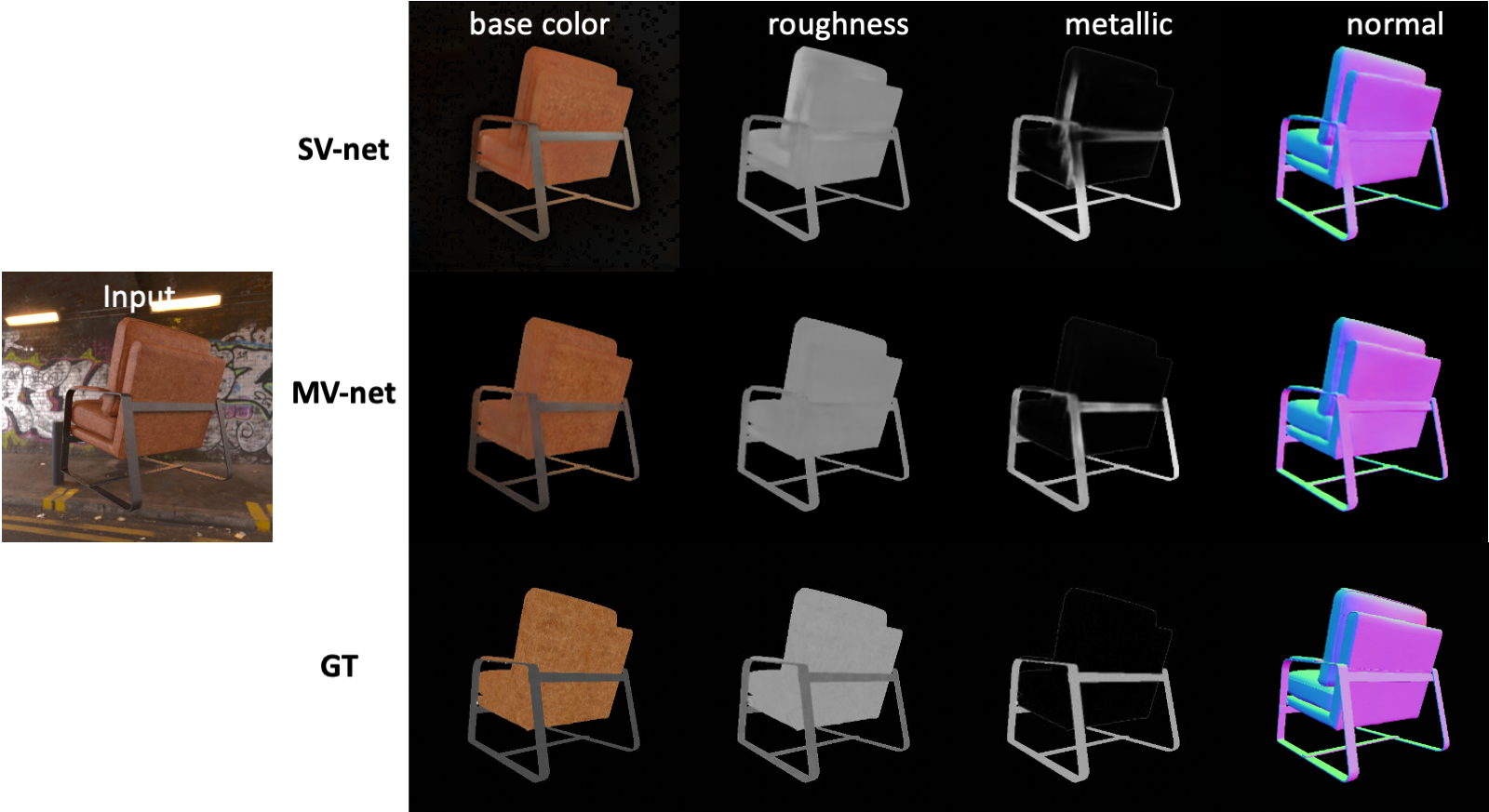}
	\includegraphics[width=0.45\linewidth]{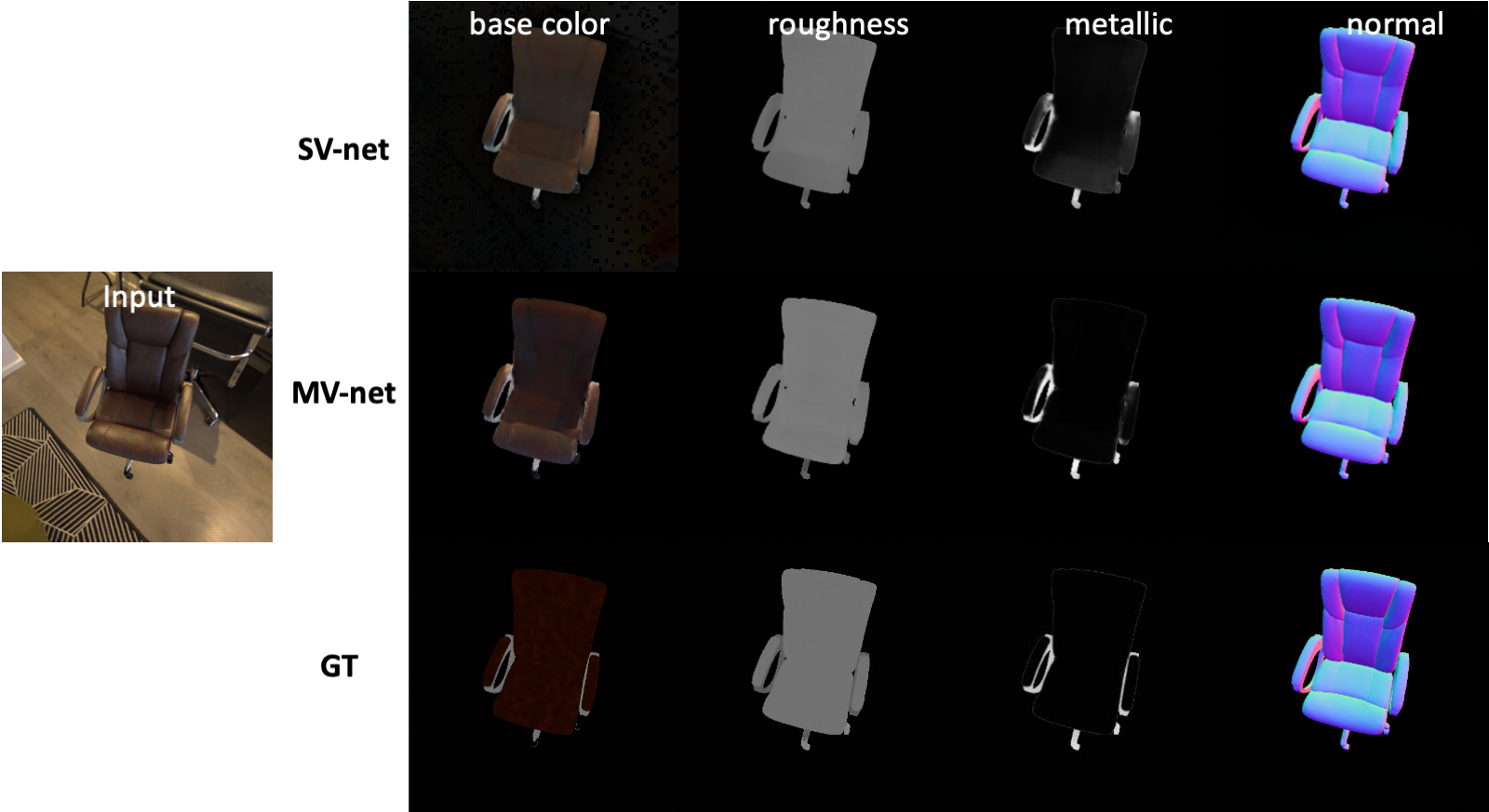}
	\caption{\textbf{Qualitative material estimation results for single-view (SV-net) and multi-view (MV-net) networks.} We show estimated SV-BRDF properties (base color, roughness, metallicness, surface normals) for each input view of an object compared to the ground truth.}
	\label{fig:material-qual}
    \vspace{-5pt}
\end{figure*}

\subsection{Material Prediction}\label{materials}
To date, there are not many available datasets tailored to the material prediction task. Most publicly available datasets with large collections of 3D objects~\cite{chang2015shapenet, choi2016large, fu20203d} do not contain physically-accurate reflectance parameters that can be used for physically-based rendering to generate photorealistic images. Datasets like PhotoShape~\cite{park2018photoshape} do contain such parameters but are limited to a single category. In contrast, the realistic 3D models in \dset are artist-created and have highly varied shapes and SV-BRDFs. We leverage this unique property to derive a benchmark for material prediction with large amounts of photorealistic synthetic data. We also present a simple baseline approach for both single- and multi-view material estimation of complex geometries.

\medskip
\noindent \textbf{Method}
To evaluate single-view and multi-view material prediction and establish a baseline approach, we use a U-Net-based model with a ResNet-34 backbone to estimate SV-BRDFs from a single viewpoint. The U-Net has a common encoder that takes an RGB image as input and has a multi-head decoder to output each component of the SV-BRDF separately. 
Inspired by recent networks in~\cite{bi2020deep, deschaintre2019flexible}, we align images from multiple viewpoints by projection using depth maps, and bundle the original image and projected image pairs as input data to enable an analogous approach for the multi-view network. We reuse the single-view architecture for the multi-view network and use global max pooling to handle an arbitrary number of input images. Similar to~\cite{deschaintre2018single}, we utilize a differentiable rendering layer to render the flash illuminated ground truth and compare it to similarly rendered images from our predictions to better regularize the network and guide the training process. Ground truth material maps are used for direct supervision.

Our model takes as input 256x256 rendered images. For training, we randomly subsample 40 views on the icosphere for each object. In the case of the multi-view network, for each reference view we select its immediate 4 adjacent views as neighboring views. We use mean squared error as the loss function for base color, roughness, metallicness, surface normal and render losses. Each network is trained for 17 epochs using the AdamW optimizer~\cite{loshchilov2017decoupled} with a learning rate of 1e\nobreakdash-3 and weight decay of 1e\nobreakdash-4.
\begin{figure}[t]
    \centering
    \includegraphics[width=1.0\linewidth]{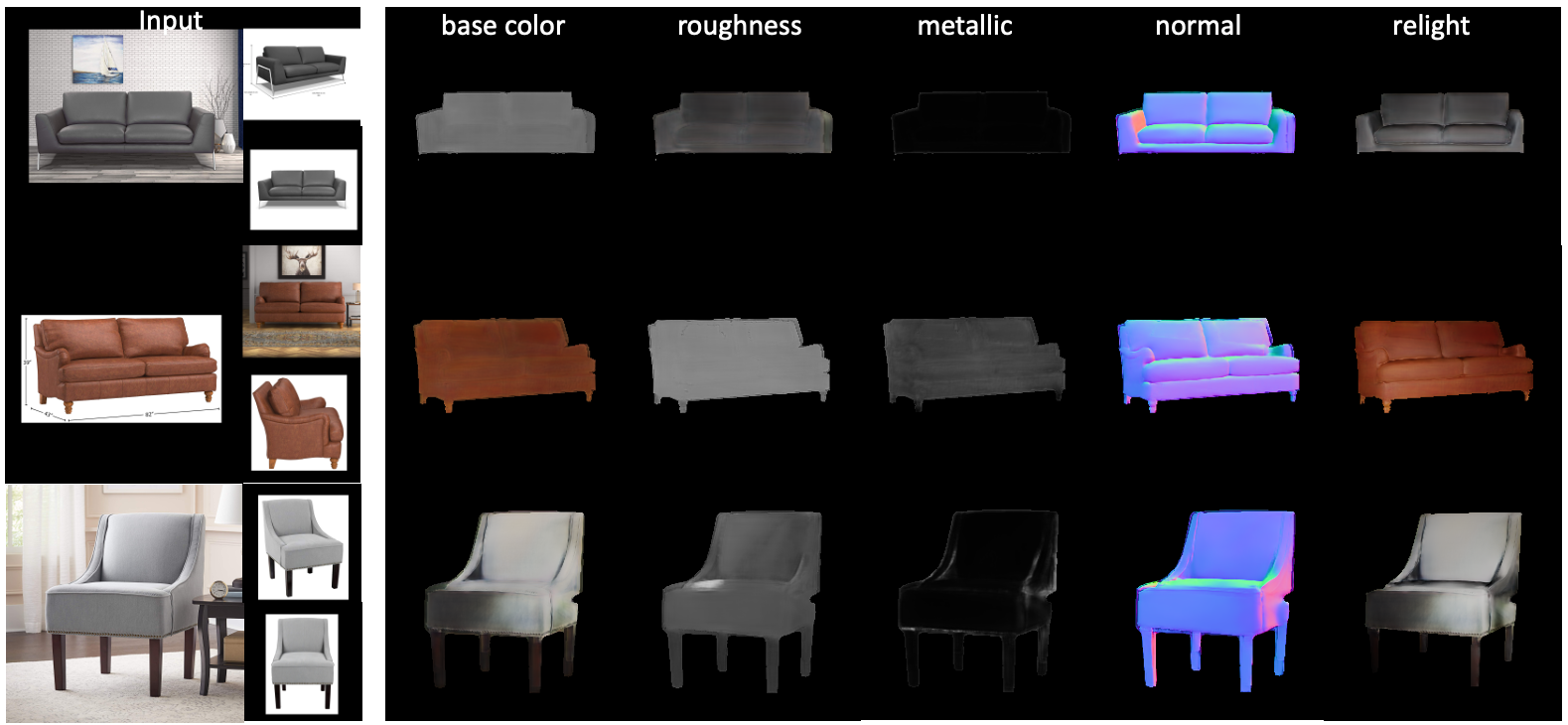}
    \vspace{-5pt}
    \caption{\textbf{Qualitative multi-view material estimation results on real catalog images.} Each of the multiple views is aligned to the reference view using the catalog image pose annotations.}
    \vspace{-10pt}
    \label{fig:material-real}
\end{figure}

\medskip
\noindent \textbf{Results}
Results for the single-view network (SV-net) and multi-view network (MV-net) can be found in Table~\ref{tab:material}.
The multi-view network has better performance compared to single-view network in terms of the base color, roughness, metallicness, and surface normal prediction tasks. The multi-view network is especially better at predicting properties that affect view-dependent specular components like roughness and metallicness.

We also run an ablation study on our multi-view network without using 3D structure to align neighboring views to reference view (denoted as \emph{MV-net: no projection}). First, we observe that even without 3D structure-based alignment, the network still outperforms the single-view network on roughness and metallic predictions. Comparing to the multi-view network, which uses 3D structure-based alignment, we can see structure information leads to better performance for all parameters. 
We show some qualitative results from the test set in Figure~\ref{fig:material-qual}. 

As a focus of \dset is enabling real-world transfer, we also test our multi-view network on catalog images of objects from the test set using the pose annotations gathered by the methodology in Section~\ref{sec:dataset}, and use the inferred material parameters to relight the object (Figure~\ref{fig:material-real}). Despite the domain gap in lighting and background, and shift from synthetic to real, our network trained on rendered images makes reasonable predictions on the real catalog images. In one case (last row), the network fails to accurately infer the true base color, likely due to the presence of self-shadow.

\begin{table}[t]
    \setlength\tabcolsep{3pt}\small
    \ra{1.1} %
	\centering
	\begin{tabular}{lccc}
        \toprule
		   &  SV-net & MV-net (no proj.) & MV-net \\
		\midrule
		Base Color ($\downarrow$) & 0.129  & 0.132 & \textbf{0.127}\\
		Roughness ($\downarrow$) &  0.163 & 0.155 & \textbf{0.129}\\
		Metallicness ($\downarrow$) &  0.170 & 0.167 &   \textbf{0.162} \\
		Normals ($\uparrow$) &  0.970 &  0.949 &  \textbf{0.976}  \\
		Render ($\downarrow$) & 0.096 & 0.090 & \textbf{0.086} \\
	\bottomrule
	\end{tabular}
	\caption{\textbf{\dset material estimation results for the single-view, multi-view, and multi-view network without projection (MV-net no proj.) ablation.} Base color, roughness, metallicness and rendering loss are measured using RMSE (lower is better) - normal similarity is measured using cosine similarity (higher is better).}
	\label{tab:material}
	\vspace{-5pt}
\end{table}

\subsection{Multi-View Cross-Domain Object Retrieval}\label{mvr}
Merging the available catalog images and 3D models in \dset, we derive a novel benchmark for object retrieval with the unique ability to measure the robustness of algorithms with respect to viewpoint changes.
Specifically, we leverage the renderings described in Section~\ref{sec:dataset}, with known azimuth and elevation, to provide more diverse views and scenes for training deep metric learning (DML) algorithms. We also use these renderings to evaluate the retrieval performance with respect to a large gallery of catalog images from \dset.
This new benchmark is very challenging because the rendered images have complex and cluttered indoor backgrounds (compared to the cleaner catalog images) and display products with viewpoints that are not typically present in the catalog images.
These two sources of images are indeed two separate image domains, making the test scenario a multi-view cross-domain retrieval task.

\medskip
\noindent \textbf{Method}
To compare the performance of state-of-the-art DML methods on our multi-view cross-domain retrieval benchmark, we use PyTorch Metric Learning~\cite{pytorchmetriclearning}
implementations that cover the main approaches to DML: NormSoftmax~\cite{zhai2018classification}
(classification-based), ProxyNCA~\cite{Movshovitz-Attias_2017_ICCV}
(proxy-based)
 and Contrastive, TripletMargin, NTXent~\cite{chen2020simple} and Multi-similarity~\cite{wang2019multi}
 (tuple-based).
 We leveraged the Powerful Benchmarker framework~\cite{powerfulbenchmarker}
 to run fair and controlled comparisons as in~\cite{musgrave2020metric}, including Bayesian hyperparameter optimization.

We opted for a ResNet-50~\cite{He_2016_CVPR} backbone, projected it to 128D after a LayerNorm~\cite{ba2016layer} layer, did not freeze the BatchNorm parameters and added an image padding transformation to obtain undistorted square images before resizing to 256x256.
We used batches of 256 samples with 4 samples per class, except for NormSoftmax and ProxyNCA where we obtained better results with a batch size of 32 and 1 sample per class.
After hyperparameter optimization, we trained all losses for 1000 epochs and chose the best epoch based on the validation Recall@1 metric, computing it only every other epoch.

Importantly, whereas catalog and rendered images in the training set are balanced (188K vs 111K), classes with and without renderings are not (4K vs. 45K).
Balancing them in each batch proved necessary to obtain good performance: not only do we want to exploit the novel viewpoints and scenes provided by the renderings to improve the retrieval performance, but there are otherwise simply not sufficiently many negative pairs of rendered images being sampled.

\medskip
\noindent \textbf{Results}
As shown in Table~\ref{tab:mvr-state-of-the-art}, the ResNet-50 baseline trained on ImageNet largely fails at the task (Recall@1 of 5\%).
This confirms the challenging nature of our novel benchmark.
DML is thus key to obtain significant improvements.
In our experiments, NormSoftmax, ProxyNCA and Contrastive performed better ($\approx\!29\%$) than the Multi-similarity, NTXent or TripletMargin losses ($\approx\!23\%$), a gap which was not apparent
in other datasets, and is not as large when using cleaner catalog images as queries.
Moreover, it is worth noting that the overall performance on \dset is significantly lower than for existing common benchmarks (see Table~\ref{tab:retrieval_benchmarks}).
This confirms their likely saturation~\cite{musgrave2020metric}, the value in new and more challenging retrieval tasks, and the need for novel metric learning approaches to handle the large scale and unique properties of our new benchmark.

\begin{figure}[t]
    \centering
    \includegraphics[width=.95\linewidth,height=.55\linewidth]{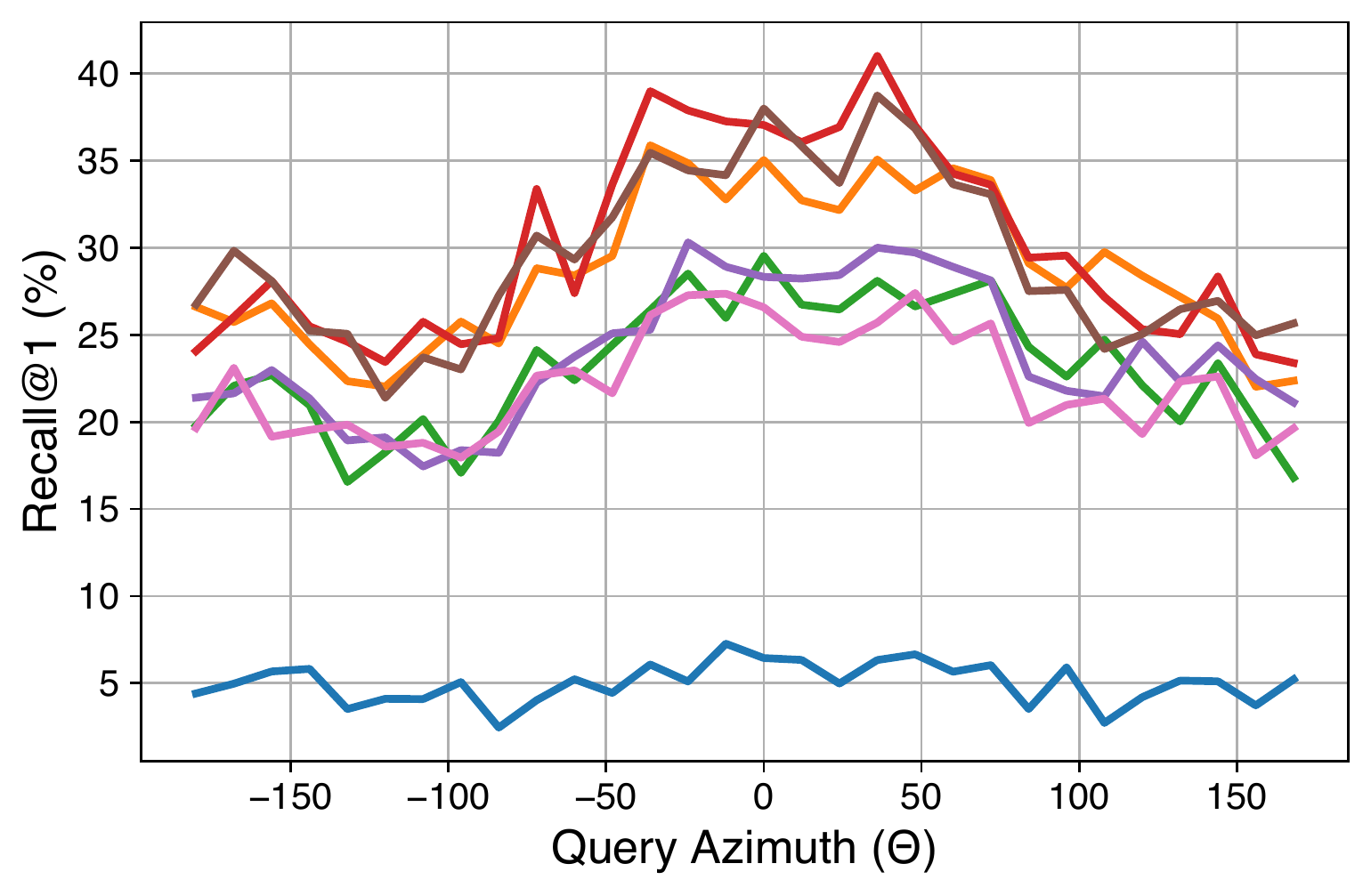}
    \includegraphics[width=.95\linewidth,height=.70\linewidth]{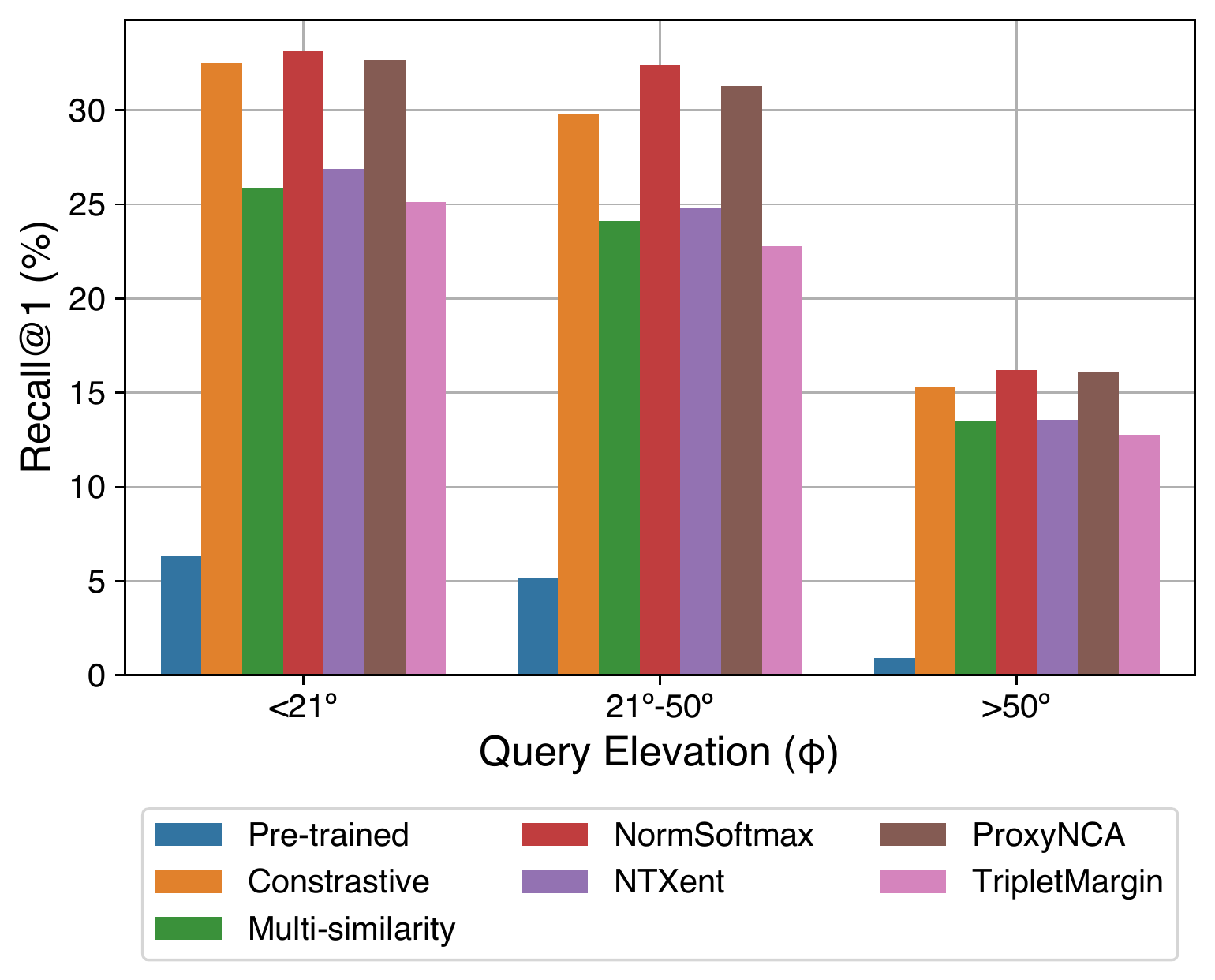}
    \vspace*{-2mm}
    \caption{\textbf{Recall@1 as a function of the azimuth and elevation of the product view.}
    For all methods, retrieval performance degrades rapidly beyond azimuth $|\theta|>75^\circ$ and elevation  $\varphi>50^\circ$.}
    \vspace{-5pt}
    \label{fig:mvr_azimuth}
\end{figure}

Further, the azimuth ($\theta$) and elevation ($\varphi$) angles available for rendered test queries allow us to measure how performance degrades as these parameters diverge from typical product viewpoints in \dset's catalog images.
Figure~\ref{fig:mvr_azimuth} highlights two main regimes for both azimuth and elevation: azimuths beyond $|\theta|\!=\!75^\circ$ and elevations above $\varphi\!=\!50^\circ$ are significantly more challenging to match, consistently for all approaches. Closing this gap is an interesting direction of future research on DML for multi-view object retrieval. For one, the current losses do not explicitly model the geometric information in training data.

\begin{table}[]
    \setlength\tabcolsep{4pt}
    \ra{1.1} %
    \centering \small
    \begin{tabular}{lrrrrr}
    \toprule
                   & \multicolumn{4}{c}{Rendered images} & Catalog \\ 
     Recall@k (\%) & $k\!=\!1$ & $k\!=\!2$ & $k\!=\!4$ & $k\!=\!8$ & $k\!=\!1$\\
    \midrule
      Pre-trained &           5.0 &           8.1 &          11.4 &          15.3 &          18.0\\
     Constrastive &          28.6 &          38.3 &          48.9 &          59.1 &          \bf{39.7} \\
 Multi-similarity &          23.1 &          32.2 &          41.9 &          52.1 &          38.0\\
      NormSoftmax &     \bf{30.0} &     \bf{40.3} &     \bf{50.2} &          60.0 &          35.5\\
           NTXent &          23.9 &          33.0 &          42.6 &          52.0 &          37.5\\
         ProxyNCA &          29.4 &          39.5 &          50.0 &     \bf{60.1} &          35.6\\
    TripletMargin &          22.1 &          31.1 &          41.3 &          51.9 &          36.9\\
    \bottomrule
    \end{tabular}
    \caption{\textbf{Test performance of state-of-the-art deep metric learning methods on the \dset retrieval benchmark.} Retrieving products from rendered images highlights performance gaps that are not as apparent when using catalog images.}
    \vspace{-5pt}
    \label{tab:mvr-state-of-the-art}
\end{table}

\section{Conclusion}
In this work we introduced \dset, a new dataset to help bridge the gap between real and synthetic 3D worlds. We demonstrated that the set of real-world derived 3D models in \dset are a challenging test set for ShapeNet-trained 3D reconstruction approaches, and that both view- and canonical-space methods do not generalize well to \dset meshes despite sampling them from the same distribution of training classes. We also trained both single-view and multi-view networks for SV-BRDF material estimation of complex, real-world geometries - a task that is uniquely enabled by the nature of our 3D dataset. We found that incorporating multiple views leads to more accurate disentanglement of SV-BRDF properties. Finally, joining the larger set of products images with synthetic renders from \dset 3D models, we proposed a challenging multi-view retrieval task that alleviates some of the limitations in diversity and structure of existing datasets, which are close to performance saturation.
The 3D models in \dset allowed us to exploit novel viewpoints and scenes during training and benchmark the performance of deep metric learning algorithms with respect to the azimuth and elevation of query images.

While not considered in this work, the large amounts of text annotations (product descriptions and keywords) and non-rigid products (apparel, home linens) enable a wide array of possible language and vision tasks, such as predicting styles, patterns, captions or keywords from product images. Furthermore, the 3D objects in \dset correspond to items that naturally occur in a home, and have associated object weight and dimensions. This can benefit robotics research and support simulations of manipulation and navigation.

\medskip
\noindent \textbf{Acknowledgements} We thank Pietro Perona and Frederic Devernay. This work was funded in part by an NSF GRFP (\#1752814) and the Amazon-BAIR Commons Program.

\clearpage

{\small
\bibliographystyle{ieee_fullname}
\bibliography{egbib}
}

\newpage
\appendix

\section{Dataset Properties}

\begin{figure}[t]
    \centering
    \vspace{10pt}
    \includegraphics[width=0.2\textwidth]{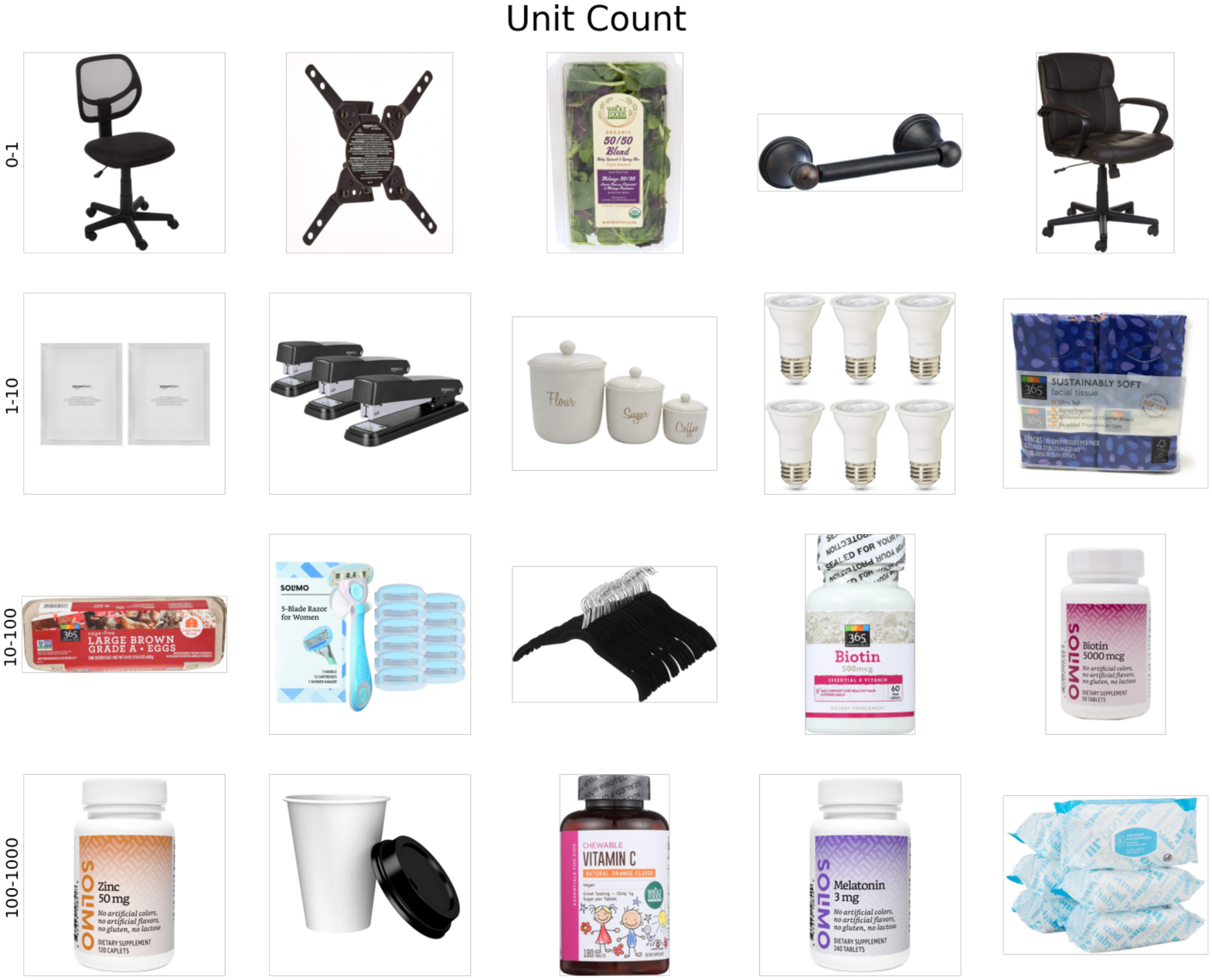} 
    \hspace{0.2cm}
    \includegraphics[width=0.2\textwidth]{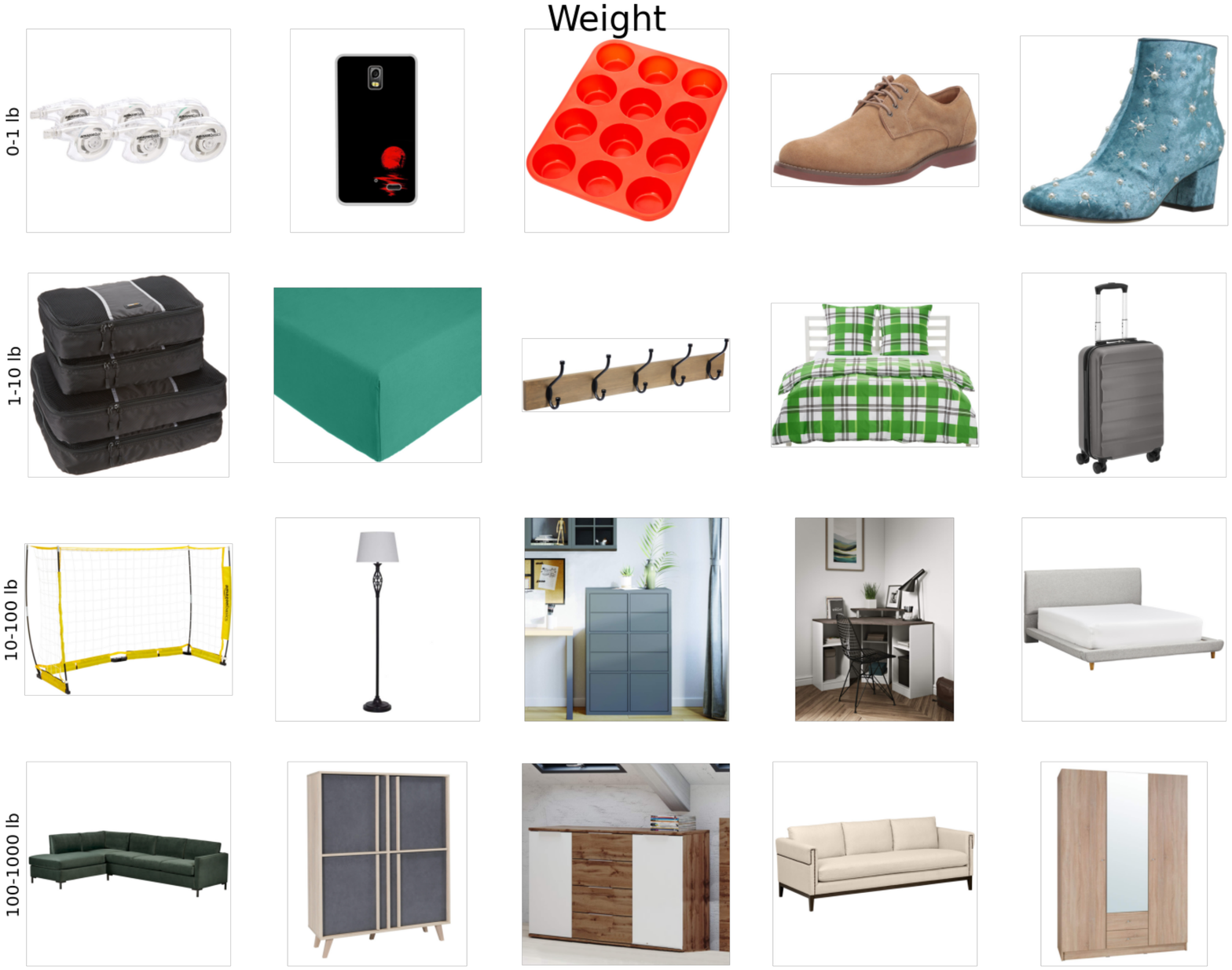}
    \caption{\textbf{Metadata visualization.} Catalog image samples for ``unit count" and ``weight" metadata attributes. For each product we show its \textit{main} image.}
    \label{fig:supp-metadata}
\end{figure}

\medskip
\noindent \textbf{Metadata Visualization}
We visualize additional products according to their metadata attributes in Figure~\ref{fig:supp-metadata}. For each attribute (unit count and weight), each row depicts products that fall into that label (for categorical attributes) or bin (for continuous-valued attributes). For continuous-valued attributes, products within the same row are ordered from lowest to highest.

\section{Dataset Organization}
In this work we used \dset to derive benchmarks for different tasks such as 3D reconstruction, material estimation, and multi-view retrieval. Table~\ref{tab:train-org} outlines how each data subset in \dset is used at both train and test time. ``No-BG Renders" refers to the white-background rendered dataset of \dset objects described in Section 4.1 of the main text, while ``BG Renders" refers to the Material Estimation Dataset described in Section 3 of the main text.

\section{6DOF Pose Optimization} \label{sec:pose}
\medskip
\noindent \textbf{Instance Masks}
We use instance masks generated both from MaskRCNN~\cite{maskRCNN} trained on LVIS~\cite{gupta2019lvis} as well as PointRend~\cite{kirillov2020pointrend} trained on COCO~\cite{lin2014microsoft}. Both model checkpoints were obtained from the Detectron2 repository\footnote{https://github.com/facebookresearch/detectron2}. We kept predicted masks from all categories with a confidence score greater than $0.1$.

\medskip
\noindent \textbf{Pose Optimization}
For each instance mask, we initialize $24$ different runs with random rotations and optimize $\textbf{R}$ and $\textbf{T}$ for $1,000$ steps using the Adam optimizer with a learning rate of 1e-2. We parameterize the rotation matrix using the symmetric orthogonalization procedure of~\cite{levinson2020analysis}. At the end of the $24$ runs, we pick the pose that has the lowest loss and validate its correctness via a human check.

\section{Single-View 3D Reconstruction Evaluation}

\medskip
\noindent \textbf{View-Space Evaluations}
 Since view-space predictions can be scaled and $z$-translated simultaneously while preserving the projection onto the camera, we must solve the depth-ambiguity to align the predicted and ground truth (GT) meshes for benchmarking. We use known camera extrinsics to transform the GT mesh into view-space, and align it with the predicted mesh by solving for a Chamfer-distance minimizing depth. In practice, we normalize average vertex depths for each mesh independently and then search through $51$ candidate depths. 
We compare the predicted and GT meshes after alignment following the evaluation protocol in \cite{fan2017point, gkioxari2019mesh} and report Chamfer distance and Absolute Normal Consistency.
Since Chamfer varies with the scale of the mesh, we follow \cite{fan2017point, gkioxari2019mesh} and scale the meshes such that the longest edge of the GT mesh bounding box is of length 10.

\begin{table}[t]
    \setlength\tabcolsep{4pt}
    \ra{1.1} %
    \centering \small
    \begin{tabular}{clccc}
    \toprule
             &    & 3D Recon. & Material Est.  & Retrieval   \\
    \midrule
     \parbox[t]{2mm}{\multirow{3}{*}{\rotatebox[origin=c]{90}{Train}}}
      & No-BG Renders    &      &          & \\
      & BG Renders    &        &  \checkmark   & \checkmark \\
      & Catalog Images    &        &    & \checkmark \\
    \midrule
    \parbox[t]{2mm}{\multirow{3}{*}{\rotatebox[origin=c]{90}{Test}}}
      & No-BG Renders    &   \checkmark    &          & \\
      & BG Renders    &        &  \checkmark   & \checkmark \\
      & Catalog Images    &        &  \checkmark   & \checkmark \\
    \bottomrule
    \end{tabular}
    \caption{\textbf{\dset data subsets used in each benchmarking experiment at both train and test time.} ``BG" and ``No-BG" Renders refer to renderings from 3D models in \dset with and without backgrounds.}
    \vspace{-5pt}
    \label{tab:train-org}
\end{table}

\medskip
\noindent \textbf{Canonical-Space Evaluations}
To evaluate shapes predicted in canonical space, we must first align them with the GT shapes. Relying on cross-category semantic alignment of models in both ShapeNet and \dset, we use a single (manually-set) rotation-alignment for the entire data. We then solve for relative translation and scale, which remain inherently ambiguous, to minimize the Chamfer distance between the two meshes. In practice, we search over a $3^4$ grid of candidate scale/translation after mean-centering (to vertex centroid) and re-scaling (via standard deviation of vertex distances to centroid) the two meshes independently. Note that R2N2~\cite{choy20163d} predicts a voxel grid so we convert it to a mesh for the purposes of benchmarking. Marching Cubes~\cite{lorensen1987marching} is one such way to achieve this, however, we follow the more efficient and easily batched protocol of ~\cite{gkioxari2019mesh} that replaces every occupied voxel with a cube, merges vertices, and removes internal faces.

\begin{figure*}
    \centering
    \includegraphics[width=0.9\textwidth]{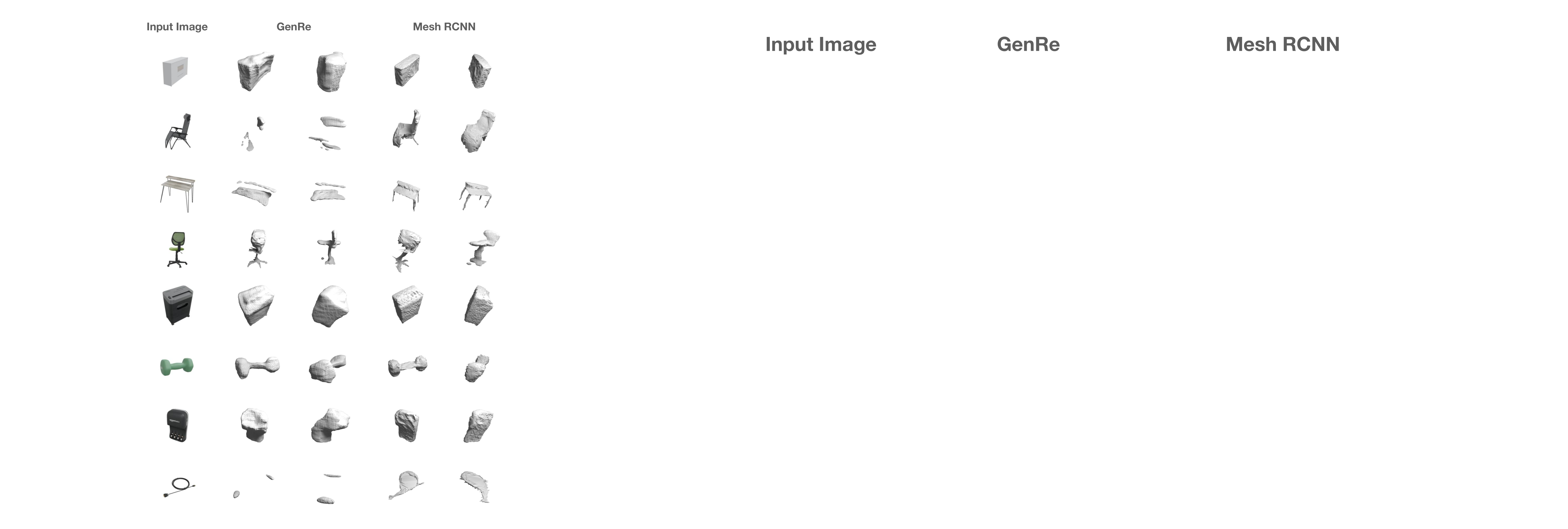}
    \caption{\textbf{Additional qualitative 3D reconstruction results for GenRe and Mesh R-CNN.} The first four rows display \dset objects that overlap with common ShapeNet categories, such as cabinet, chair and table. The bottom four rows display reconstructions for \dset objects from categories not represented in ShapeNet.}
    \label{fig:supp-qual3d}
\end{figure*}

\medskip
\noindent \textbf{Additional Qualitative Examples}
We show additional reconstructions of \dset objects from the best performing single-view 3D reconstruction method, Mesh R-CNN, and the method that claims to be category-agnostic, GenRe in Figure~\ref{fig:supp-qual3d}. We find that both methods generally fail to reconstruct objects with thin structures, but that the failure occurs in different ways for each method. GenRe simply does not reconstruct them, whereas Mesh R-CNN produces a reconstruction that qualitatively appears more like the 3D convex hull of the object.

\medskip
\noindent \textbf{Quantitative Evaluation on Test Split}
As the focus of this work was to measure the domain gap of ShapeNet-trained 3D reconstruction methods to \dset, we evaluated reconstruction performance for all 3D models in \dset that came from ShapeNet categories. To facilitate future research that may want to \textit{train} 3D reconstruction methods using \dset and compare to ShapeNet trained methods, we generate a train/val/test split (80\%/10\%/10\%) that we release along with the dataset. We also report the performance (averaged across all categories) of each method on this test split in Table~\ref{tab:test-set-recon}.

\begin{table}[]
    \setlength\tabcolsep{4pt}
    \ra{1.1} %
    \centering \small
    \begin{tabular}{lcc}
    \toprule
                 & Chamfer ($\downarrow$) & Abs. Normal Consistency ($\uparrow$) \\
    \midrule
      3D R2N2~\cite{choy20163d}    &           1.97 &           0.55 \\
      OccNets~\cite{mescheder2019occupancy}    &           1.19 &           0.70 \\
      GenRe~\cite{zhang2018genre}     &           1.61 &           0.66 \\
      Mesh R-CNN~\cite{meshrcnn} &           0.82 &           0.62 \\
    \bottomrule
    \end{tabular}
    \caption{\textbf{3D reconstruction on \dset test split.} Chamfer distance and absolute normal consistency averaged across all categories}
    \vspace{-15pt}
    \label{tab:test-set-recon}
\end{table}

\section{Material Estimation}
\medskip
\noindent \textbf{Dataset Curation}
We use the Material Estimation Dataset outlined in the main text, omitting object with transparencies, resulting in 7,679 models. We split the 3D models into a non-overlapping train/test set of 6,897 and 782 models, respectively. To test generalization to new lighting conditions, we reserve 10 out of 108 HDRI environment maps for the test set only. 

\begin{figure}[t]
  \centering
  \includegraphics[width=0.98\linewidth]{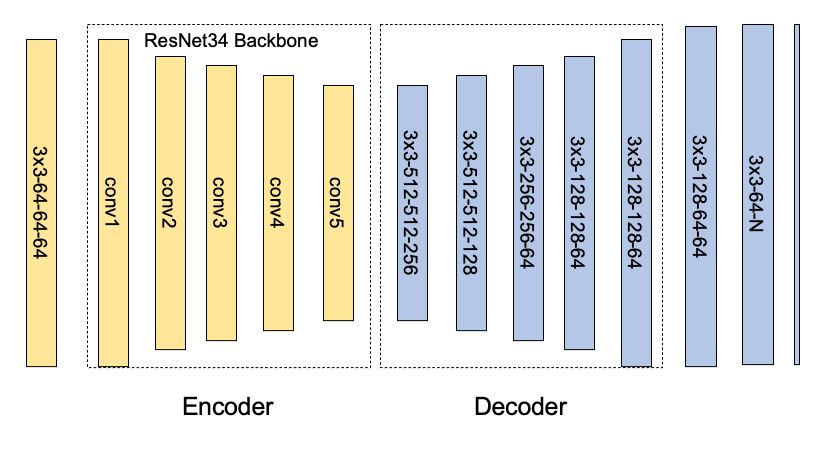}
  \includegraphics[width=0.98\linewidth]{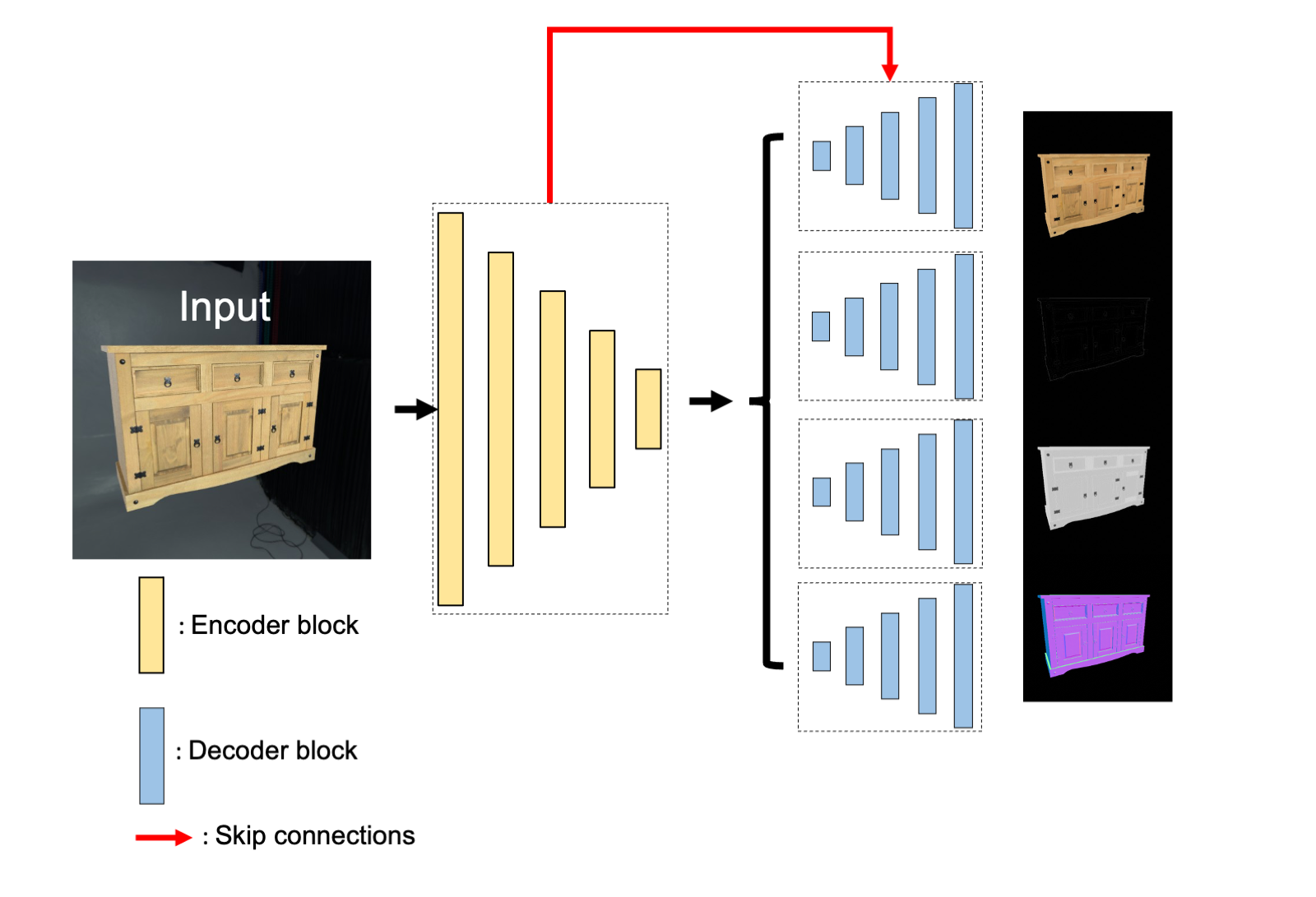} \\
  \vspace{10pt}
  \includegraphics[width=0.98\linewidth]{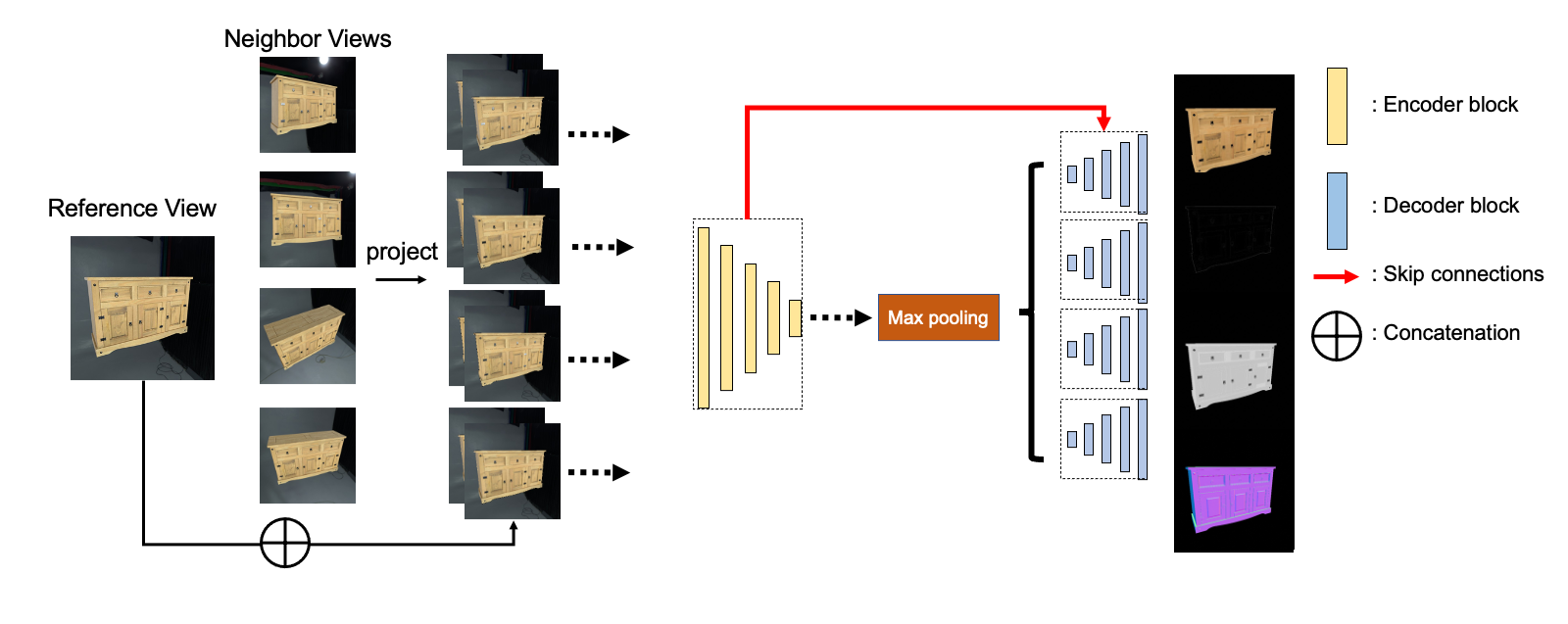}
  \caption{\textbf{Top:} Encoder-decoder architecture. Encoder uses a ResNet-34 (conv1-conv5) backbone. \textit{KxK-N-M-X} denotes a double convolution block of \textit{KxK} filter, \textit{N} input channels, \textit{M} intermediate channels, and \textit{X} output channels. We use BatchNorm and leaky ReLU. \textbf{Middle:} Single-view baseline. \textbf{Bottom:} Multi-view baseline. Given a reference view, neighboring views are selected and projected to the reference view and passed as input to the network.
  }
  \label{fig:single-figure} 
\end{figure}

\medskip
\noindent \textbf{Network Details}
A visualization of the single-view material estimation network (SV-net) and multi-view network (MV-net) can be found in Figure~\ref{fig:single-figure} (middle and bottom, respectively). The MV-net uses camera poses to establish pixel-level correspondences. Given a pixel $p$ in one viewpoint with image coordinate $x$ and depth $z$, the image coordinate $x'$ of the its corresponding pixel in another viewpoint can be computed as $
x' = KRK^{-1} x + Kt/z$, where $K$ is the camera intrinsic matrix, $R$ and $t$ are the rotation and translation between the two viewpoints. 
Some pixels in one view can be occluded and hence not visible in other views. These can be determined by using a depth-based occlusion test. We fill these pixels using values from the reference view.

\begin{figure}[h]
    \centering
    \includegraphics[width=1.0\linewidth]{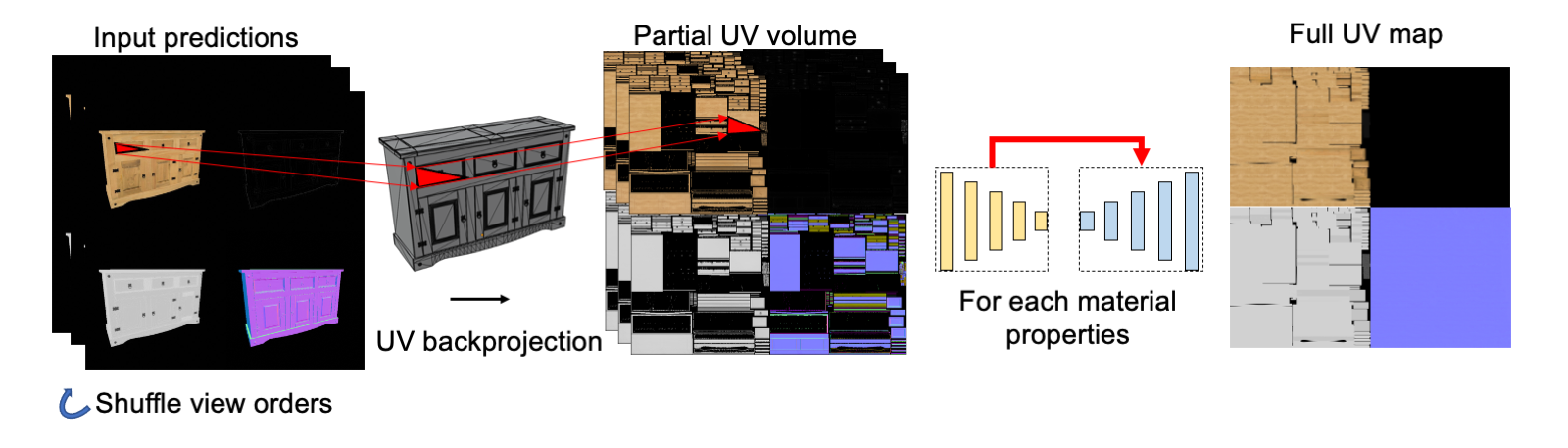}
    \caption{\textbf{Pipeline for full UV map prediction.} Per-view predictions are back projected onto the UV and fed to an encoder-decoder network for smoothing and aggregation. }
    \label{fig:reconstruction}
\end{figure}

\medskip
\noindent \textbf{Full Texture Map Reconstruction}
Using per-view predicted material maps, we can also generate a full textured PBR model. This is achieved by back-projecting the predicted material maps to the UV domain and using a learned network to aggregate and smooth the predictions. Figure~\ref{fig:reconstruction} shows the full UV map reconstruction pipeline. 

\section{Multi-View Cross-Domain Object Retrieval}

\medskip
\noindent \textbf{Dataset Curation}\label{curation}
For this task, we used product type annotations to focus only on rigid objects, removing items such as garments, home linens, and some accessories (cellphone accessories, animal harnesses, cables, etc.). 
As the set of products beyond just those with 3D models are likely to contain near-duplicates (i.e. different sizes of the same shoe), we then applied a hierarchical Union-Find algorithm for near-duplicate detection and product grouping, based on shared imagery as a heuristic.
We considered near-duplicates as correct matches of a single instance and thus assigned a unique \emph{instance id} to all near-duplicate listings.
Product groups are sets of such instances that are from product lines that may share design details, materials, patterns and thus may have common images (close-up detail of fabric, fact sheet image, ...).
Consequently, we ensured that all instances in a group are assigned to the same data split (train, val or test).
The val and test sets contain only instances with 3D models and, while their catalog images compose their respective target set (val-target and test-target), we use rendered images as queries (val-query and test-query).

\medskip
\noindent \textbf{Dataset Curation Statistics}
The hierarchical Union-Find algorithm yielded $29,988$ groups of $50,756$ instances, of which $1,334$ have 3D models ($5,683$ instances).
We then sampled groups accounting for $836$ instances with 3D models for the \emph{test} set, using their combined $4,313$ catalog images as \emph{test-target} images and sampled 8 rendered views for each of the environment maps as \emph{test-query} images.
Again, we sampled groups with $854$ of the remaining instances with 3D models for the \emph{validation} set, using catalog images ($4,707$) as \emph{val-target} and 8 rendered images per envmap as \emph{val-query}.
We used the rest of the instances as the \emph{train} set: $49,066$ instances ($3,993$ with 3D model), $187,912$ catalog images and $110,928$ rendered images ($298,840$ total).

\medskip
\noindent \textbf{Implementation Details}
Train images are pre-processed by the following: square padding, resize to 256x256, random resized crop of 227x227 (scale between $0.16$ and $1$ and ratio  between $0.75$ and $1.33$) and random horizontal flip ($p=0.5$). Test images are padding to square, resized to 256x256 and center cropped to 227x227. 

The trunk used is ResNet-50 pre-trained on ImageNet, leading to a 2048D vector. We did not freeze these weights, including the BatchNorm parameters. The embedding module is a LayerNorm normalization followed by linear projection to 128D embedding. The batch sampler is class-balanced and domain-balanced. Domain-balancing is implemented as a hierarchical sampling: we first sample $N$ classes with rendered images and $N$ classes without, then sample $K$ images for each class among all available images for the class, leading to batches of $2NK$ samples. For NormSoftmax and ProxyNCA we used batches of 32 samples, 1 sample per class, 16 classes with rendered images and 16 without. For all other methods we used batches of 256 samples, 4 samples per class, 64 classes with rendered images and 64 classes without.

One epoch consists of 200 batches sampled from the above procedure. We trained the models for 1000 epochs, using early stopping when no improvements on the validation accuracy are found for over 250 consecutive epochs. The epoch for testing is selected based on the maximum validation Recall@1 for validation. We used 8 workers for data loading on a AWS p3.8xlarge instance (32 cores, 4 Nvidia V100 GPUs). As~\cite{musgrave2020metric}, we did not use tuple mining within a batch. The embeddings are normalized before indexing and querying. We used Recall@1 using \emph{val-query} rendered images as queries against \emph{val-target} catalog images as the validation metric, computing it at every even epoch.
RMSProp with a learning rate $1e\!-\!6$, weight decay $1e\!-\!4$, and momentum 0.9 is used to optimize both the trunk and mebedding layers. For metric losses, where applicable, the learning rate is optimized via hyperparameter optimization.

\begin{table*}[th]
    \setlength\tabcolsep{4pt}
    \ra{1.1} %
    \centering \small
    \begin{tabular}{lrrrrrrr}
    \toprule
                 Loss &  Recall@1 (\%) &  Recall@2 (\%) &  Recall@4 (\%) &  Recall@8 (\%) &  MAP (\%) &  MAP@R (\%) &  R-Precision (\%) \\
    \midrule
          Pre-trained &          4.97 &          8.10 &         11.41 &         15.30 &     7.69 &       2.27 &             3.44 \\
         Constrastive &         28.56 &         38.34 &         48.85 &         59.10 &    31.19 &      \bf{14.16} &            \bf{19.19} \\
     Multi-similarity &         23.12 &         32.24 &         41.86 &         52.13 &    26.77 &      11.72 &            16.29 \\
          NormSoftmax &    \bf{30.02} &     \bf{40.32} &.  \bf{50.19} &         59.96 &  \bf{32.61} &      14.03 &            18.76 \\
               NTXent &         23.86 &         33.04 &         42.59 &         51.98 &    27.00 &      12.05 &            16.51 \\
             ProxyNCA &         29.36 &         39.47 &         50.05 &     \bf{60.11} &    32.38 &      14.05 &            19.00 \\
        TripletMargin &         22.15 &         31.10 &         41.32 &         51.90 &    25.80 &      10.87 &            15.41 \\
    \bottomrule
    \end{tabular}
    \caption{\textbf{Test metrics for the ABO-MVR benchmark, using rendered images as queries}. 
    The gallery images are derived from catalog images and contain classes from the train and test classes.}
    \label{tab:mvr-testquery}
\end{table*}

\begin{table*}[th]
    \setlength\tabcolsep{4pt}
    \ra{1.1} %
    \centering \small
    \begin{tabular}{lrrrrrrr}
    \toprule
                 Loss &  Recall@1 (\%) &  Recall@2 (\%) &  Recall@4 (\%) &  Recall@8 (\%) &  MAP (\%) &  MAP@R (\%) &  R-Precision (\%) \\
    \midrule
          Pre-trained &         17.99 &         23.93 &         31.72 &         38.65 &    22.57 &       6.99 &             9.55 \\
         Constrastive &     \bf{39.67} &.   \bf{52.21} &    \bf{64.41} &    \bf{71.64} &    \bf{42.96} &      \bf{22.52} &            \bf{28.07} \\
     Multi-similarity &         38.05 &         50.06 &         61.79 &         68.17 &    40.87 &      21.06 &            26.32 \\
          NormSoftmax &         35.50 &         46.70 &         57.38 &         64.78 &    38.07 &      18.63 &            23.42 \\
               NTXent &         37.51 &         49.34 &         61.37 &         69.23 &    40.12 &      20.03 &            25.32 \\
             ProxyNCA &         35.64 &         46.53 &         57.36 &         65.06 &    38.50 &      18.81 &            23.65 \\
        TripletMargin &         36.87 &         48.34 &         60.98 &         69.44 &    40.03 &      19.94 &            25.46 \\
    \bottomrule
    \end{tabular}
    \caption{\textbf{Test metrics for the ABO-MVR benchmark, using catalog images as queries}. The gallery images are derived from catalog images and contain classes from the train and test classes.}
    \label{tab:mvr-testtarget}
\end{table*}

\par Hyperparameters are estimated by 30 runs of Bayesian hyperparameter optimization for 100 epochs, using the implementation of Powerful-Benchmarker, the epoch is chosen on the full run after optimization:

\begin{itemize}
    \setlength\itemsep{0em}
    \item Contrastive. Pos margin:~$0.5287$; Neg margin:~$1.0523$; Epoch:~$614$
    \item Multi-similarity. Alpha:~$0.0240$; Beta:~$49.1918$; Base:~$0.5567$; Epoch:~$76$
    \item NormSoftmax. Temperature:~$0.0776$; Metric loss learning rate:~$0.0013$; Epoch:~$448$
    \item NTXent. Temperature:~$0.0665$; Epoch:~$114$
    \item ProxyNCA. Softmax scale:~$5.5915$; Metric loss learning rate:~$0.0010$; Epoch:~$566$
    \item Triplet. Margin:~$0.0743$; Epoch:~$706$
\end{itemize}

\medskip
\noindent \textbf{Additional Metrics}
Tables~\ref{tab:mvr-testquery} and~\ref{tab:mvr-testtarget} provide a more complete version of Table 5 from the main manuscript.
In addition to Recall@k, we also report mean average precision (MAP), mean average precision at R (MAP@R) and R-Precision as desribed and implemented by~\cite{musgrave2020metric}.
These metrics provide additional insights compared to recall, as they give higher value to retrieval results with correct rankings and having as many correct results as possible in the first ranks.

In Table~\ref{tab:mvr-testquery}, we compare the results when using rendered images of test classes as queries, against the union of catalog images of test classes and catalog images of train classes.
In Table~\ref{tab:mvr-testtarget}, we instead compare the results when using catalog images of test classes as queries, against the same union of catalog images of test classes and catalog images of train classes.
Naturally, the image used as query is discarded from its own search results before computing the metrics.
Still, as we can see, these is a significant gap between retrieving from rendered queries compared to catalog images. This highlights the difficulty of this new benchmark.

\medskip
\noindent\textbf{Qualitative Results}
In Figure~\ref{fig:mvr_qualitative} we show qualitative results for a few queries (low elevation, mid elevation, high elevation), showing some success and failure cases of NormSoftmax, ProxyNCA and Contrastive.

\newlength{\mvrknnwidth}
\setlength{\mvrknnwidth}{0.98\textwidth}

\begin{figure*}
\centering
    \includegraphics[width=\mvrknnwidth]{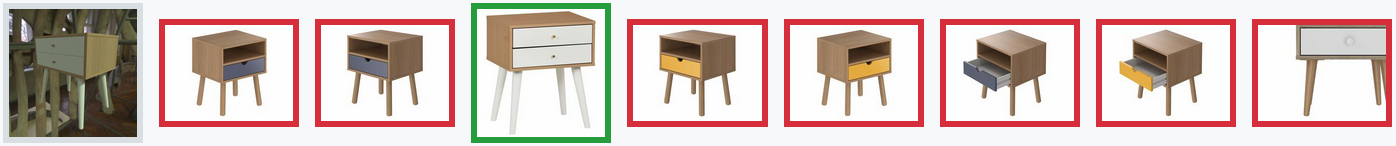}
    \includegraphics[width=\mvrknnwidth]{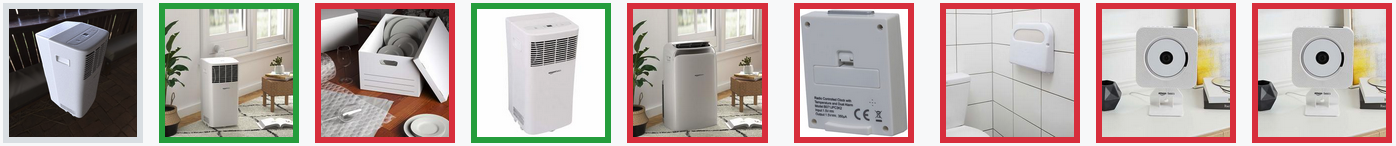}
    \includegraphics[width=\mvrknnwidth]{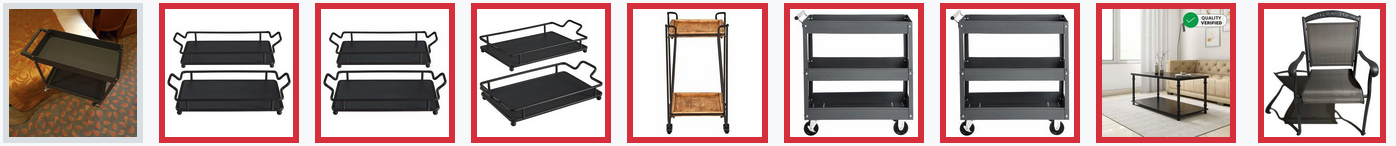}
    \\ \vspace*{5mm} \hrule{} \vspace*{5mm}
    \includegraphics[width=\mvrknnwidth]{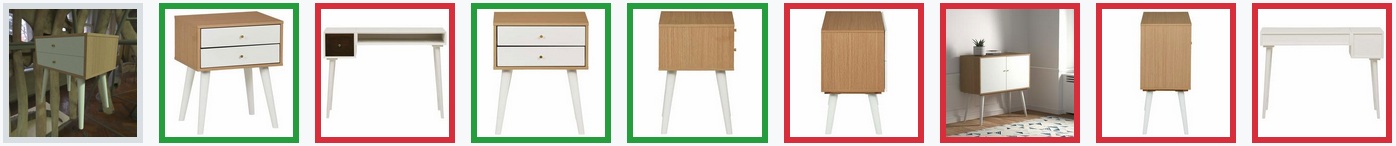}
    \includegraphics[width=\mvrknnwidth]{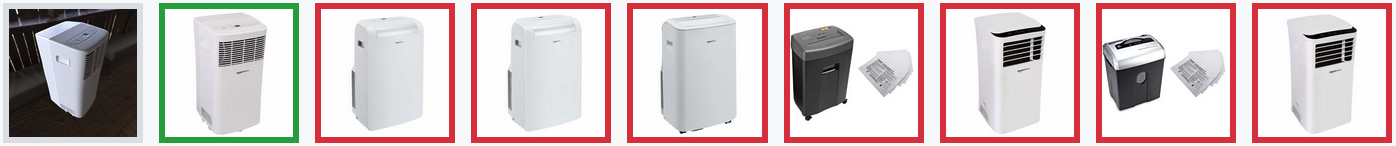}
    \includegraphics[width=\mvrknnwidth]{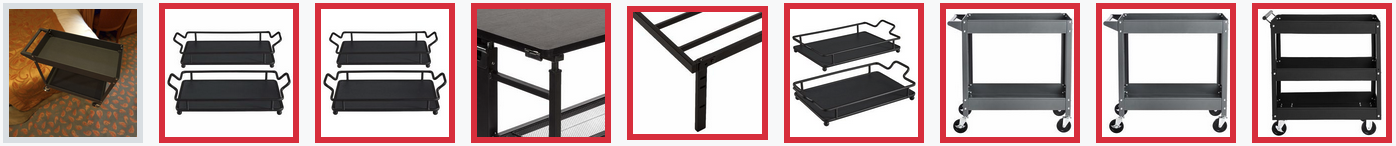}
    \\ \vspace*{5mm} \hrule{} \vspace*{5mm}
    \includegraphics[width=\mvrknnwidth]{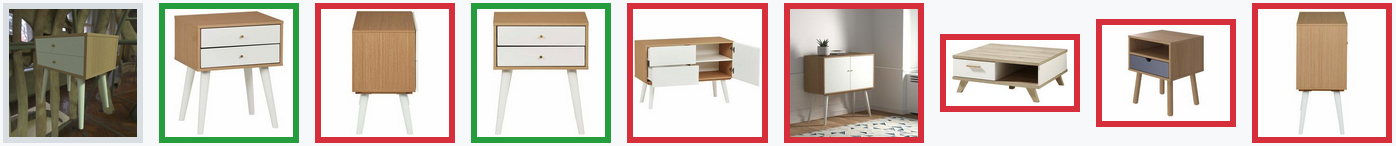}
    \includegraphics[width=\mvrknnwidth]{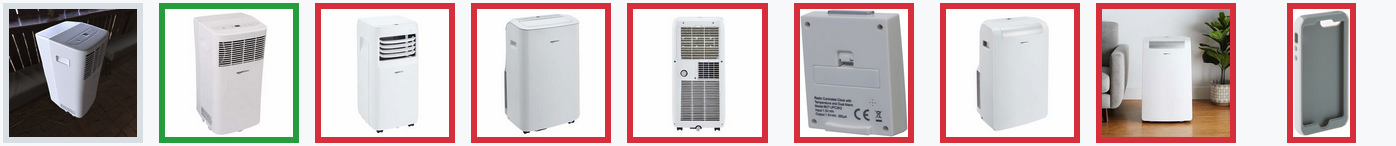}
    \includegraphics[width=\mvrknnwidth]{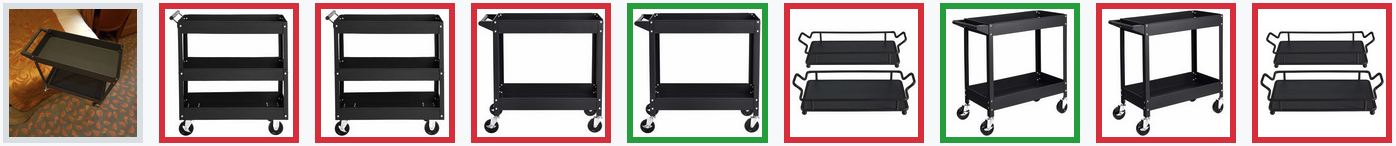}
    
\caption{\textbf{Qualitative retrieval results for low, medium, and high elevation products.} The leftmost column shows the query image, the other 8 columns show the top-8 results, highlighted in green if correct and red if incorrect. The top 3 rows are results of Contrastive, the middle 3 are of NormSoftmax and the bottom 3 are ProxyNCA. Each group of 3 rows have the same queries: one of low elevation (side table), one of mid-elevation (air conditioner), one of high elevation (cart).}
\label{fig:mvr_qualitative}
\end{figure*}

\end{document}